\documentclass[conf]{new-aiaa}
\usepackage[utf8]{inputenc}
\usepackage{textcomp}

\usepackage{graphicx}
\usepackage{amsmath}
\usepackage{siunitx}
\usepackage{longtable,tabularx}
\usepackage{subcaption}
\usepackage{xcolor}
\usepackage{todonotes}
\usepackage{graphicx}
\usepackage{subcaption}

\usepackage{amsthm}
\theoremstyle{definition}
\newtheorem{defn}{Definition}

\newcommand{\R}{\mathbb{R}}

\title{Learning-based methods to model small body gravity fields for proximity operations: Safety and Robustness}

\author{Daniel Neamati \footnote{Current affiliation: Ph.D. Student, Aeronautics and Astronautics, and AIAA Student Member. \href{mailto:dneamati@stanford.edu}{dneamati@stanford.edu}, 734-904-8477.}}
\affil{California Institute of Technology, Pasadena, CA, 91125}
\author{Yashwanth Kumar Nakka \footnote{Postdoctoral Fellow, Jet Propulsion Laboratory, California Institute of Technology, and AIAA Student Member. \href{mailto:ynakka@jpl.nasa.gov}{ynakka@jpl.nasa.gov}}}
\affil{California Institute of Technology, Pasadena, CA, 91125}
\author{Soon-Jo Chung \footnote{Bren Professor of Aerospace and Control and Dynamical Systems, Division of Engineering and Applied Science, California Institute of Technology, MC 105-50, and an Associate Fellow of AIAA. \href{mailto:sjchung@caltech.edu}{sjchung@caltech.edu}}}
\affil{California Institute of Technology, Pasadena, CA, 91125}

\begin{document}

\maketitle

\begin{abstract}
Accurate gravity field models are essential for safe proximity operations around small bodies. State-of-the-art techniques use spherical harmonics or high-fidelity polyhedron shape models. Unfortunately, these techniques can become inaccurate near the surface of the small body or have high computational costs, especially for binary or heterogeneous small bodies. New learning-based techniques do not encode a predefined structure and are more versatile. In exchange for versatility, learning-based techniques can be less robust outside the training data domain. In deployment, the spacecraft trajectory is the primary source of dynamics data. Therefore, the training data domain should include spacecraft trajectories to accurately evaluate the learned model's safety and robustness. We have developed a novel method for learning-based gravity models that directly uses the spacecraft's past trajectories. We further introduce a method to evaluate the safety and robustness of learning-based techniques via comparing accuracy within and outside of the training domain. We demonstrate this safety and robustness method for two learning-based frameworks: Gaussian processes and neural networks. Along with the detailed analysis provided, we empirically establish the need for robustness verification of learned gravity models when used for proximity operations.
% Using our method, we establish the need for robustness verification of learned gravity models.  
% \todo[inline]{180 words}
\end{abstract}

% \todo[inline]{TO-DO: Re check AIAA format after text and figures are added}
% AIAA figures are "Fig. #" in general and "Figure #" at the start of the sentence
% AIAA tables are "Table #" everywhere
% AIAA equations are "Eq. (#)" in general and "Equation #" at the start of the sentence. "Eqs." for multiple.

% \todo[inline]{TO-DO: Should we use capital or lowercase for framework names (i.e., ``Gaussian Process Model" or ``gaussian process model" or ``Gaussian process model"?}
% Based on other works, I decided to go with "Gaussian process" and "neural network." I use "--- model" to refer to a gravity model and I use "---" or "--- framework" to refer to the approach.

\section*{Nomenclature}
{\renewcommand\arraystretch{1.0}
\noindent\begin{longtable*}{@{}l @{\quad=\quad} l@{}}
$a$ & Dimensionless acceleration \\
$\tilde{a}_{\text{semi-major}}$ & Normalized, dimensionless semi-major axis \\
$a_{\text{semi-major}}$ & Semi-major axis (\si{\metre}) \\
$\epsilon$ & Fractional error \\
$\tilde{J}_n$ & The $n$-th normalized zonal spherical harmonic \\
$\mu$ & The gravitational parameter (\si{\cubic\metre\per\square\second}) \\
$N$ & Number of samples from the training trajectory \\
$\sigma_{s, a}$ & State or acceleration uncertainty standard deviation, respectively \\
$\tau_{\text{Kepl.}}$ & Instantaneous, normalized Keplerian period (\si{\second}) \\
$u_n$ & Contribution to the specific gravitational potential from the $\tilde{J}_n$ term (\si{\square\metre\per\square\second}) \\
$u_H$ & The value of the potential at the Hill radius (\si{\square\metre\per\square\second}) \\
$x$ & Dimensionless position 
\end{longtable*}}

\section{Introduction}
Recent missions to small bodies (e.g., \textit{Rosetta}, \textit{Hayabusa 2}, and \textit{OSIRIS-REx}) have reshaped our understanding of small bodies and inspired new, more-capable future missions. New missions, such as \textit{Lucy}, \textit{DART}, and \textit{Psyche}, suggest that there remains an increasing demand for new small body missions. Still, large uncertainties in the gravity field around small bodies make proximity operations for these missions challenging. Flying with an inaccurate dynamics model incurs substantial risks for proximity maneuvers and landing, including spacecraft collision with the small body. Ideally, we could build the gravity model directly from the small body's density distribution, but an instrument that can measure the internal density distribution to provide the requisite data does not exist. State-of-the-art methods assume a small body is of constant density and use the body's shape to develop a gravity model \cite{Werner1997ExteriorCastalia, Scheeres2016OrbitalOrbiters, Hockman2017DesignBodies}. However, during the planetary accretion process, small bodies can form as binaries (such as 67P/Churyumov-Gerasimenko and 486958 Arrokoth) or rubble piles (such as 101955~Bennu) \cite{Scheeres2016OrbitalOrbiters, Scheeres2020HeterogeneousBennu}. Binary and rubble pile small bodies are composed of materials with diverse densities. For these heterogeneous small bodies, the constant density assumption does not hold. We need new methods to estimate a gravity model to safely reach targets of heterogeneous density.

\begin{figure}
    \centering
    \includegraphics[width = 0.8 \textwidth]{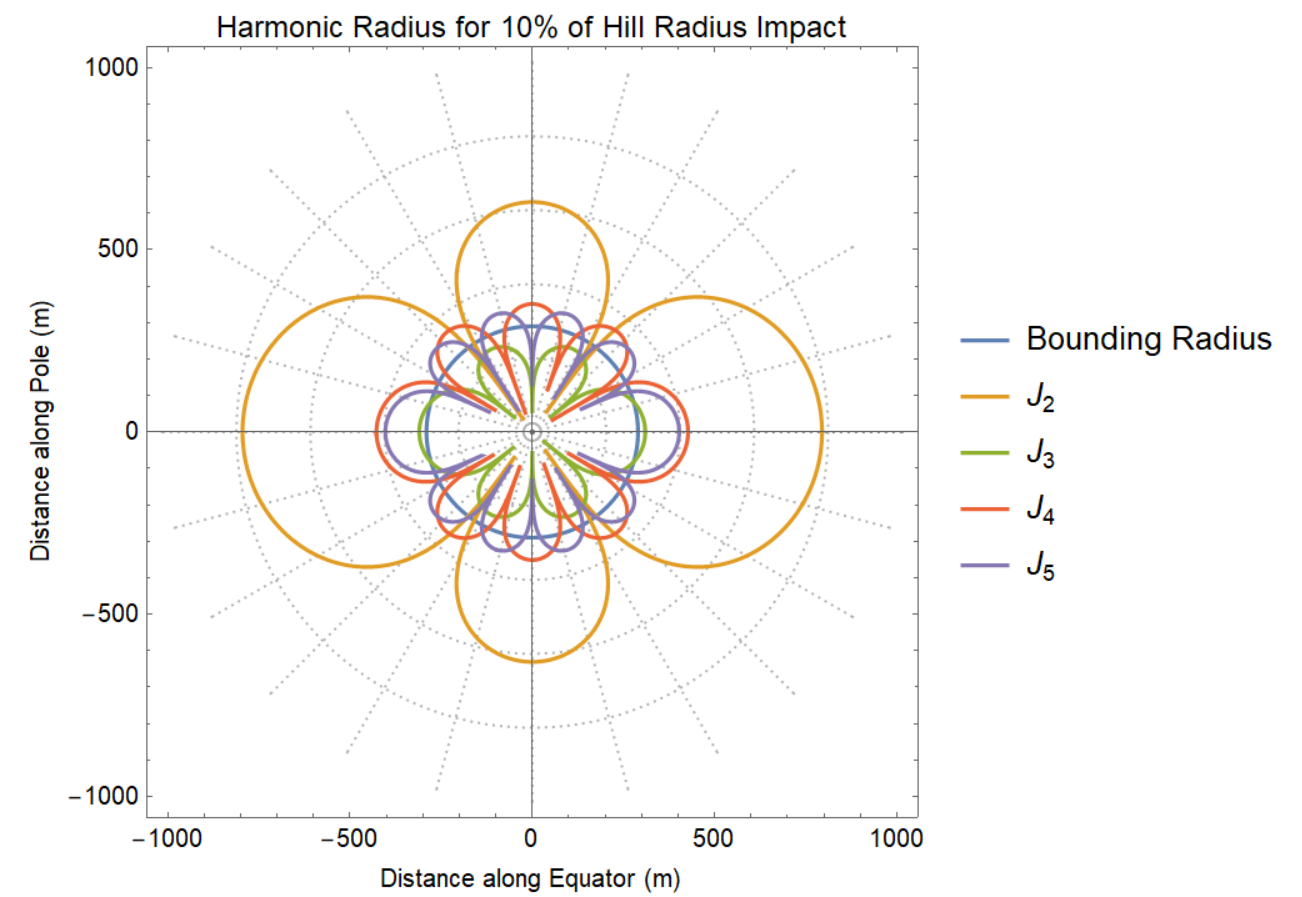}
    \caption{Radius as a function of co-latitude where the specific potential of the $n$-th normalized zonal spherical harmonic term is 10\% of the specific potential at the Hill Radius for the asteroid 101955~Bennu. Within each contour, the impact of the term increases as $|\frac{u_n}{u_H}|^{n + 1}$ (where $u_n$ is the contribution to the potential from the $n$-th harmonic and $u_H$ is the value of the potential at the Hill Radius). The spherical harmonics are not valid within the bounding radius (shown as the blue circle). For 101955~Bennu and similar small bodies, the harmonics are only appreciable in close proximity to the surface. Adapted from \cite{Neamati2021NewFieldsFixed}.}
    \label{fig:jn_impact}
\end{figure}

For proximity maneuvers, the gravitational force from the small body overwhelms other contributions such as solar radiation pressure or solar gravity. While the point mass term overwhelmingly dominates, the higher-order terms are appreciable near the surface (Fig.~\ref{fig:jn_impact}). The irregular shape and heterogeneous density distributions of the small body combined are responsible for this complex gravity field near the surface \cite{Scheeres2016OrbitalOrbiters, Scheeres2020HeterogeneousBennu, Hockman}. The challenge centers on a cyclic problem: we need a high-fidelity gravity model to plan a safe trajectory, but we simultaneously need to visit the small body on a safe trajectory to estimate the high-fidelity gravity model. Polyhedron and mascon gravity models generally use a high-fidelity shape model of the body assuming a constant density to build a gravity model \cite{Werner1997ExteriorCastalia, Scheeres2016OrbitalOrbiters, Hockman2017DesignBodies}. Additional multi-density polyhedron and finite element type methods relax the constant density assumption. This group of methods can represent objects of multiple densities but at the cost of increased computational load. Still, these multi-density gravity models have difficulties near the density boundaries \cite{Takahashi2013GravityBodies, Takahashi2014MorphologyBodies}. Spherical harmonics models are well-suited for planet-sized or moon-sized objects but, they are inadequate for small body landing maneuvers since the series diverges within the minimum bounding sphere (also known as the Brillouin Sphere). On the other hand, recent learning-based methods require processed gravitational data from previous missions. To achieve good performance, they necessitate training data that covers the whole orbital domain of interest. They densely sample the predictions of a prespecified reference gravity model (such as a polyhedron model) to generate this training data \cite{Song2019FastApplication, Martin2020Applications21-393}. There are two issues to extending this method to an unvisited small body. First, if the prespecified gravity model has faulty assumptions, the learning-based method will directly replicate the incorrect assumptions. Second, the learning-based methods require a prespecified gravity model, but such a model is unavailable for an unvisited small body.

\begin{figure}
    \centering
    \includegraphics[width = 0.9 \textwidth, trim={0cm 8cm 0cm 2cm}, clip]{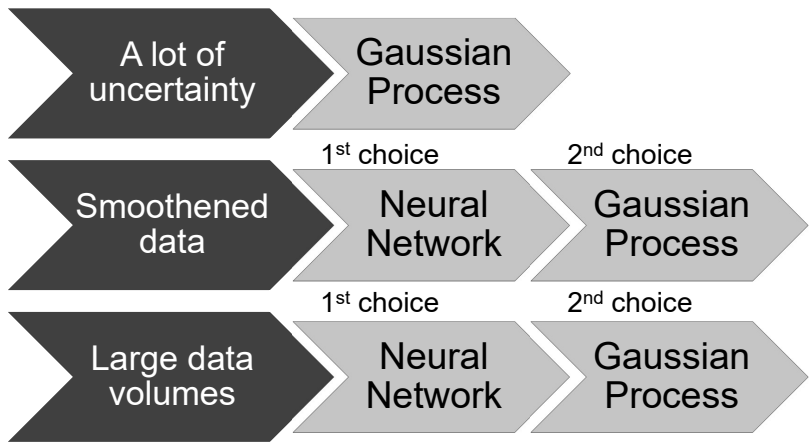}
    % {%
% \setlength{\fboxsep}{0pt}%
% \setlength{\fboxrule}{1pt}%
% \fbox{\includegraphics[width = 0.9 \textwidth, trim={0cm 8cm 0cm 2cm}, clip]{Figures/recommendations-new-grayscale.pdf}}%
% }%
    \caption{An overview of recent work from \cite{Neamati2021NewFieldsFixed}. Based on recent work, the Gaussian process model outperforms other frameworks in the case of high uncertainty. If the signal can be sufficiently filtered or smoothened, the neural network models outperform the Gaussian process models. When using large data volumes, the high computational load of the Gaussian process model is a significant limitation. All three of these parameters can change how we characterize the safety and robustness of the learning framework. As a architects for these learning-based techniques, we must identify the range of uncertainties, data smoothening, and data volumes that matches the mission demands. Adapted from \cite{Neamati2021NewFieldsFixed}. }
    \label{fig:recommendations}
\end{figure}

To overcome these challenges, we developed a novel learning-based gravity model that uses spacecraft trajectory data to ensure an unbiased estimate of the small body gravity field. With this formulation, the learned gravity model adapts to onboard data directly in the learning framework to improve the model throughout the mission. In this research, we considered two types of learning frameworks: Gaussian processes (extending off \cite{Gao2019EfficientRegression}) and neural networks with spectral normalization (adding continuity guarantees on \cite{Cheng2020Real-TimeNetworks, Song2019FastApplication, Martin2020Applications21-393}). In our earlier analysis, we observed that Gaussian processes generally outperform other frameworks in cases of moderate uncertainty \cite{Neamati2021NewFieldsFixed}. As the uncertainty declines or the data is sufficiently filtered, neural networks with spectral normalization provide more accurate gravity models and are computationally cheaper to evaluate (Fig.~\ref{fig:recommendations}). However, to the author's knowledge, there is no framework to assess the safety or robustness of a learning-based gravity model. Here, we address this gap and introduce a new method to begin evaluating the safety and robustness of learning-based gravity models.

\subsection{Safety and Robustness Characterization of Gravity Models}
Safety and robustness are broad concepts. For this paper, we focus on measuring safety and robustness via acceleration prediction accuracy. We define safety and robustness as follows.
\begin{defn}(Safety.) We characterize safety using the acceleration prediction error of an interpolation test set compared to the training data set. A trajectory is considered safe if the prediction error is small. 
\end{defn}
In characterizing safety, we seek to understand how accurately the predicted accelerations match the true accelerations during typical operations. Accurate acceleration predictions ensure that the spacecraft can maintain safe, collision-free trajectories around the small body. % To assess safety, we compare how accurately the learning framework performs on a test set interpolated within the training domain to the training data set itself.
\begin{defn}(Robustness.) We characterize robustness with the divergence of prediction error from the training domain to an extrapolation test set. We can robustly transition to a new trajectory if the error is small along the new trajectory.
\end{defn}

 For robustness, we seek to understand how the accuracy of the prediction diverges outside of the training domain. A slow divergence ensures that the spacecraft can robustly recover from trajectories that deviated from a nominal trajectory. To assess robustness, we compare how accurately the learning framework extrapolates to a test trajectory. We draw this test trajectory from the same distribution as the training data set but on a different orbit. Throughout the paper, we discuss additional ways to characterize safety and robustness for a learning-based gravity model.

\subsection{Paper Organization}
In this paper, we discuss each new method and present representative results along the way. In Section~\ref{sec:frameworks}, we contrast the Gaussian process and neural network models. We detail how to extend these gravity models for future work. In Section~\ref{sec:traj-data}, we condense our novel method for trajectory-only learned dynamics that we first presented in \cite{Neamati2021NewFieldsFixed}. We focus on our procedures for data generation, training, and testing. In Section~\ref{sec:characterization}, we discuss our novel method for safety and robustness characterization. We demonstrate this method for a single learned model before extending it to a class of learning models.

\section{Learning Frameworks \label{sec:frameworks}}
To demonstrate breadth in performance and capabilities, we considered two learning frameworks: Gaussian processes and neural networks. First, Gaussian process models would be particularly well suited for stochastic optimal control, conditional on computational efficiency and accuracy. The stochastic part of the controller incorporates covariance information of the spacecraft state and the learned dynamics % \cite{Oguri, Nakka2019TrajectorySystems}
% \cite{Oguri}
\cite{Oguri,Nakka2019TrajectorySystems, Nakka2021TrajectoryControl, Nakka2021SpacecraftLearning}. The Gaussian processes natively provide this covariance information \cite{Rasmussen2006GaussianLearning}. Moreover, prior work suggests that Gaussian process regression may be suited for gravity field estimation \cite{Gao2019EfficientRegression}. Second, the structure of the neural network enables quicker evaluations than the Gaussian processes. But, the neural network approach does not natively subsume Lipschitz continuity, nor does it estimate the uncertainty in its evaluation. Neglecting continuity and uncertainty impedes ready incorporation into stochastic optimal control frameworks. Adding spectral normalization, we can mathematically guarantee a Lipschitz continuous output \cite{Cosgrove2018SpectralExplained, Shi2019NeuralDynamics, Shi2020Neural-Swarm:Interactions}. With this research, we incorporate spectral normalization into prior work \cite{Cheng2020Real-TimeNetworks, Song2019FastApplication, Martin2020Applications21-393}. We further probe the neural network's robustness under different uncertainty thresholds. %As with the Gaussian process model, we contextualize the neural network model strategy for stochastic optimal control. 
For future safety and robustness metrics that incorporate covariance information, we suggest that future work should incorporate covariance information into the training data and use a covariance net structure, as in \cite{Russell2019MultivariateLearning}. Table~\ref{tab:LearningFrameworkVariants} illustrates the Gaussian process and neural network used in this work. Note that the Gaussian process and neural network approaches do not encompass an exhaustive set of machine learning strategies. For example, a physics informed loss function could improve performance, as in \cite{Martin2020Applications21-393}. Other architectures such as recurrent neural networks or deep sets could also be valuable in learned dynamics for gravity \cite{Shi2020Neural-Swarm:Interactions}. The design space for learning-based architectures is immense. The community needs specific metrics of performance before we can evaluate which architectures are the best performing. This paper focuses on introducing such a metric rather than optimizing any specific architecture.

\begin{table}
\caption{\label{tab:LearningFrameworkVariants} Learning framework architecture selection}
\centering
\begin{tabular}{c|p{5cm}p{5cm}}
\hline
        Parameter Types & Gaussian Process & Neural Network \\
        \hline
        Subtype Focus & Exact Gaussian process & Fully connected neural network \\
        Optimizer & Adam with constant learning rate & Adam with constant learning rate \\
        Loss Function & Exact marginal log likelihood & Mean squared error\\
        \hline
        Framework Specific & Constant mean & Spectral normalization\\
        & Radial-basis function kernel & Rectified linear unit activation\\
        & & 6 hidden layers with 80 nodes each \\
\hline
\end{tabular}
\end{table}

\section{Trajectory-only Learned Dynamics \label{sec:traj-data}}
In our previous work, we detailed a new method for trajectory-only learned dynamics of small body gravity fields \cite{Neamati2021NewFieldsFixed}. In this section, we summarize our prior work and focus on the details most relevant to safety and robustness characterization.

\subsection{Physics Engine}
\begin{figure}
    \centering
    \includegraphics[width = 0.9 \textwidth]{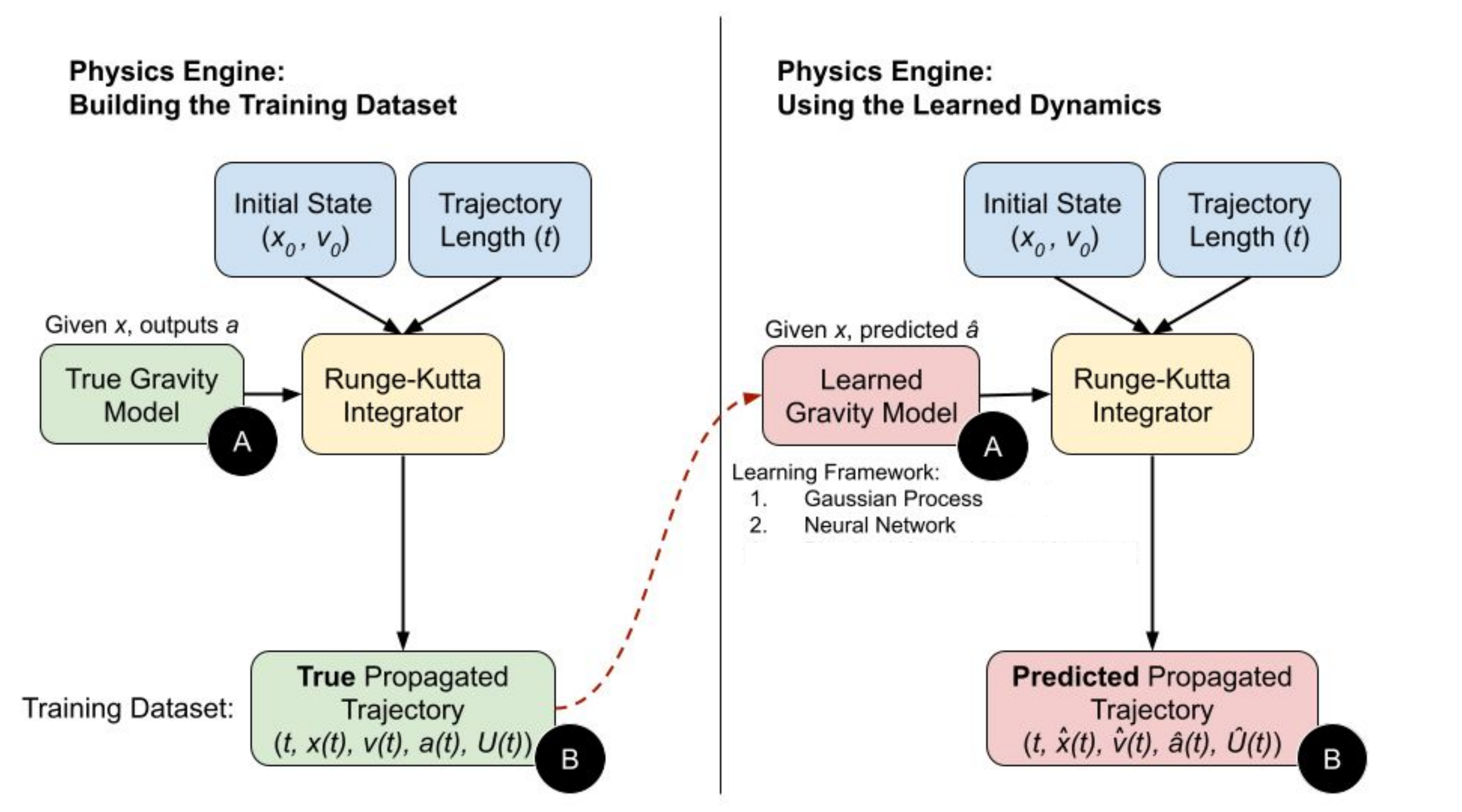}
    \caption{Diagram of the physics engine architecture. We use the reference gravity model to build the training trajectory dataset (left side and green boxes). We then apply any shuffling or noise to the trajectory (termination of the dashed red arrow). After training the learned model (dashed red arrow), we arrive at the trained, learned gravity model (red). We then use the learned gravity model to build the predicted trajectory dataset (right side and red boxes). The test point between pointwise direct model outputs is labeled ``A,'' and the test point between trajectory outputs is labeled ``B.'' Our metrics for safety and robustness focus on test point ``A,'' but future work could extend metrics to test point ``B.'' Adapted from \cite{Neamati2021NewFieldsFixed}.}
    \label{fig:simulator_architecture}
\end{figure}

In the absence of spacecraft data, the physics engine generates all the data for training, safety characterization, and robustness characterization (Fig.~\ref{fig:simulator_architecture}). We used a standard fourth-order Runge-Kutta integrator to propagate an initial state in $\R^6$ (position and velocity) over a user-defined trajectory length with a reference gravity model. In this analysis, we base the gravity model on 101955~Bennu using \cite{Scheeres2020HeterogeneousBennu} as a reference (Table~\ref{tab:bennu-grav}).   

\begin{table}
\caption{\label{tab:bennu-grav} Asteroid 101955~Bennu as a reference gravitational field. Adapted from \cite{Scheeres2020HeterogeneousBennu, Neamati2021NewFieldsFixed}.}
\centering
\begin{tabular}{c|c}
\hline
        Property & Value  \\
        \hline
        Brillouin sphere radius & 290 \si{\metre} \\
        Gravitational parameter ($\mu$) & 4.89 \si{\cubic\metre\per\square\second} \\
        $\tilde{J}_2$ & $1.93 \times 10^{-2}$ \\
        $\tilde{J}_3$ & $-1.22 \times 10^{-3}$ \\
        $\tilde{J}_4$ & $-6.50 \times 10^{-3}$ \\
        $\tilde{J}_5$ & $6.73 \times 10^{-5}$ \\
        \hline
    \end{tabular}
\end{table}

We then sample the trajectory at a user-defined, even spacing to discretize the trajectory. Lastly, we incorporate any uncertainty to the propagated trajectory to better replicate spacecraft data. The learning framework uses the state information ($S \in \R^3$) as an input and the acceleration information ($A \in \R^3$) as the output. For a data set of size $N$, the data volume is $(|S| + |A|) \times N = 6 N$. We randomly shuffle the data prior to training to prevent temporal autocorrelation during the framework training.

\subsection{Data Generation \label{subsec:datagen}}
\begin{figure}
    \centering
    \includegraphics[width=0.9 \textwidth]{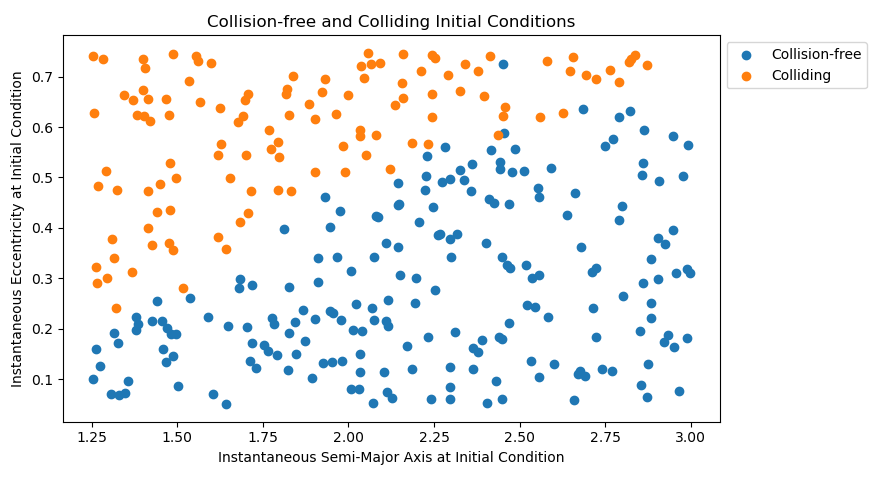}
    \caption{The instantaneous normalized Keplerian orbital elements of initial conditions for collision-free and colliding trajectories projected onto semi-major axis and eccentricity. The collision-free and colliding boundary is roughly linear when projected onto the semi-major axis and eccentricity. However, since the gravitational field is not uniform, the inclination and other orbital elements impact the collision-free and colliding boundary. A collision model that uses the shape rather than the Brillouin sphere will also change the boundary shape.}
    \label{fig:initialconditions}
\end{figure}

During the data generation process, we purposely sample trajectories with periapsis near the surface of the small body where the higher-order gravitational terms are most prominent, and the risks for inaccuracy are highest. Since we are using past spacecraft trajectory data, we only have access to collision-free trajectories. We parameterize the trajectories by the instantaneous, normalized Keplerian orbital elements. In particular, we consider a point mass representation of the small body with distances normalized to the average radius of the small body. The instantaneous, normalized Keplerian period defined in Eq.~(\ref{eq:kepl-period}) is a proxy for the time scale for the trajectory. Since the body is irregularly shaped and heterogeneous in density, the orbit will not be Keplerian, nor will the orbital period match the normalized Keplerian period. However, the instantaneous, normalized Keplerian orbital elements are particularly useful in visualizing the start of the trajectory and defining the orbital state domain.
\begin{equation}
    \tau_{\text{Kepl.}} = 2 \pi \sqrt{\frac{a_{\text{semi-major}}^3}{\mu}} \label{eq:kepl-period}
\end{equation}
Table~\ref{tab:normalize-kepl-ranges} details the normalized Keplerian orbital element ranges used in the data generation process. The longitude of the ascending node ($\Omega$), the argument of perigee ($\omega$), and the true anomaly ($\nu$) are restricted for plotting purposes but can be extended without modification. To separate the collision-free trajectories from the colliding trajectories, we propagate the orbit forward for 50 orbits as defined by the instantaneous Keplerian orbit. We use the same gravity model as before (Table~\ref{tab:bennu-grav}). %(i.e., normalized harmonics as $\Tilde{J}_2 = 1.926 \times 10^{-2}, \Tilde{J}_3 = -1.22 \times 10^{-3}, \Tilde{J}_4 = -6.499 \times 10^{-3},$ and $\Tilde{J}_5 = 6.729 \times 10^{-5}$). 
We mark a collision if a trajectory enters the body's minimum bounding sphere (i.e., the Brillouin Sphere). More advanced future work can use the small body shape model directly to determine a collision. Figure~\ref{fig:initialconditions} illustrates the collision-free and colliding boundary projected onto the semi-major axis and eccentricity. Future work can explore different collision-free and colliding boundary classification techniques.

% Normalized Harmonics 1.926e-2, -1.22e-3, -6.499e-3, 6.729e-5
% a_range: tuple = (1.25, 3.0), e_range: tuple = (0.05, 0.75),
%                  i_range: tuple = (0.0, 180.0), o_range: tuple = (0.0, 180.0),
%                  w_range: tuple = (0.0, 180.0), f_range: tuple = (0.0, 180.0)

\begin{table}
\caption{\label{tab:normalize-kepl-ranges} Instantaneous normalized Keplerian orbital element ranges for data generation}
\centering
\begin{tabular}{ccc}
\hline
        Orbital Element & Range Lower Bound & Range Upper Bound \\
        \hline
        $\tilde{a}_{\text{semi-major}}$ & 1.25 & 3 \\
        e & 0.05 & 0.75 \\
        i & $0^\circ$ & $180^\circ$ \\
        $\Omega$ & $0^\circ$ & $180^\circ$ (up to $360^\circ$) \\
        $\omega$ & $0^\circ$ & $180^\circ$ (up to $360^\circ$) \\
        $\nu$ & $0^\circ$ & $180^\circ$ (up to $360^\circ$) \\
\hline
\end{tabular}
\end{table}

\begin{figure}
    \centering
    
    \begin{subfigure}[b]{0.49\textwidth}
         \centering
         \includegraphics[width = \textwidth]{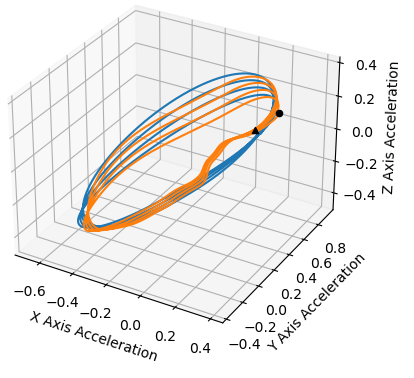}
         \caption{Gaussian Process Model}
    \end{subfigure}
    % \begin{subfigure}[b]{0.5\textwidth}
    %      \centering
    %      \includegraphics[width = \textwidth]{ResultsFigures/GP1-acceleration_l2_error_0-titlefixed.png}
    %      \caption{Error in Prediction for Training Set 2}
    % \end{subfigure}
    \begin{subfigure}[b]{0.49\textwidth}
         \centering
         \includegraphics[width = \textwidth]{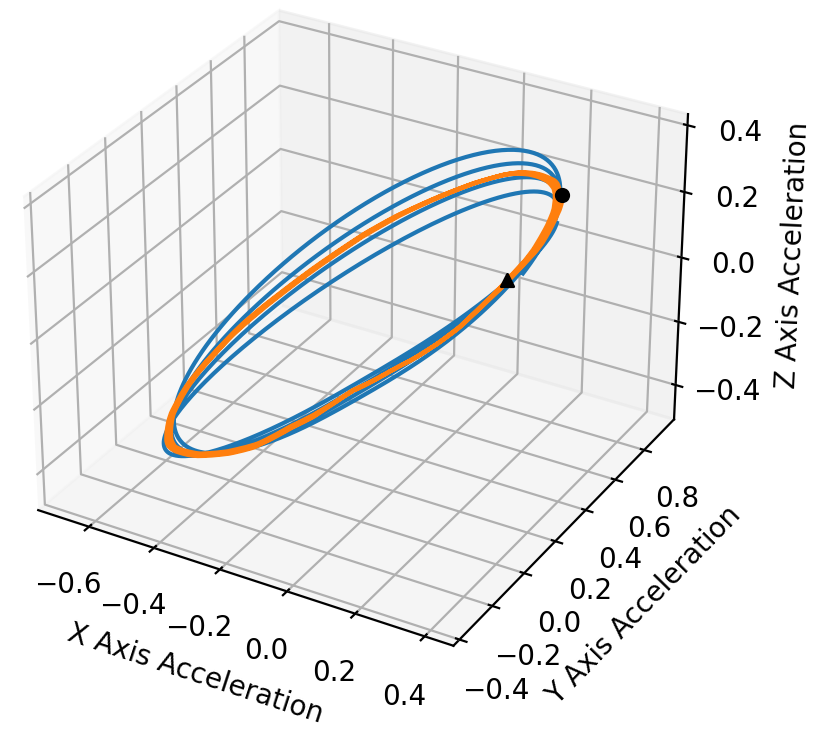}
         \caption{Neural Network Model}
    \end{subfigure}
    % \begin{subfigure}[b]{0.5\textwidth}
    %      \centering
    %      \includegraphics[width = \textwidth]{ResultsFigures/NN2-acceleration_3d_1_clean_sn.png}
    %      \caption{Acceleration Trajectories}
    % \end{subfigure}
    \caption{Example predictions using the Gaussian process model framework (a) or neural network model framework with spectral normalization (b). The blue curve is the true acceleration along the trajectory, and the orange curve is the predicted acceleration. The black circles mark the initial positions of the trajectories. The black triangles mark the final positions. Adapted from \cite{Neamati2021NewFieldsFixed}.}
    \label{fig:gpnn-accel-traj-comparison}
\end{figure}

\subsection{Learning}
We train the frameworks for a fixed number of epochs at a fixed learning rate and fixed loss function. While each of these can be relaxed, fixing these parameters provides better consistency between runs. Moreover, it also ensures a bounded training time. We find that the Gaussian process models take roughly four times as long to train as the neural network models at low data volumes. At higher data volumes, the gap grows further. Having a fixed number of epochs at a fixed learning rate ensures that we can output a learned model within a bounded time window. We can observe when the model fails to learn or hits numerical instabilities in the post-processing discussed later in this paper. The frameworks must learn the acceleration function $a$ that best maps the input states to output accelerations (Eq.~(\ref{eq:learned-dynamics})). This learning task is more difficult under uncertainty where neither the state nor acceleration is perfectly observable.
\begin{equation}
    a_{\text{data}} \approx a_{\text{learned}}(x_{\text{data}}) \approx a(x_{\text{true}}) = a_{\text{true}} \label{eq:learned-dynamics}
\end{equation}
In orbit, the spacecraft is unable to observe the true state ($x_{\text{true}}$), the true accelerations at the state ($a_{\text{true}}(x_{\text{true}})$), or the true acceleration function (i.e., the true gravity model, $a: x_{\text{true}} \in \R^3 \to a_{\text{true}} \in \R^3$). Instead, we observe the state data ($x_{\text{data}}$) and acceleration data ($a_{\text{data}}$). The learning framework finds the best function ($a_{\text{learned}}$) that maps the state data to the acceleration data. Hence, we minimize the gap between $a_{\text{data}}$ and $a_{\text{learned}}(x_{\text{data}})$. After training, we can compare the true values to the learned values. Namely, we compare $a_{\text{learned}}(x_{\text{true}})$ to $a(x_{\text{true}}) = a_{\text{true}}$. Since the problem is mapping a three dimensional state space to a three dimensional acceleration space, it is difficult to visualize the predictions. Instead we can parameterically plot the outputs along the trajectory. This changes $a(x): \R^3 \to \R^3$ to $a(x(t)): \R \to \R^3$. Figure~\ref{fig:gpnn-accel-traj-comparison} is an example of the acceleration predictions as parameteric plots.

% \missingfigure{Propagated orbit (if we need more content) \label{fig:propagated-orbit}}

% \newpage
\section{Safety and Robustness Characterization \label{sec:characterization}}

\subsection{Single instance characterization \label{subsec:singlechar}}

\begin{figure}
    \centering
    \includegraphics[width=\textwidth]{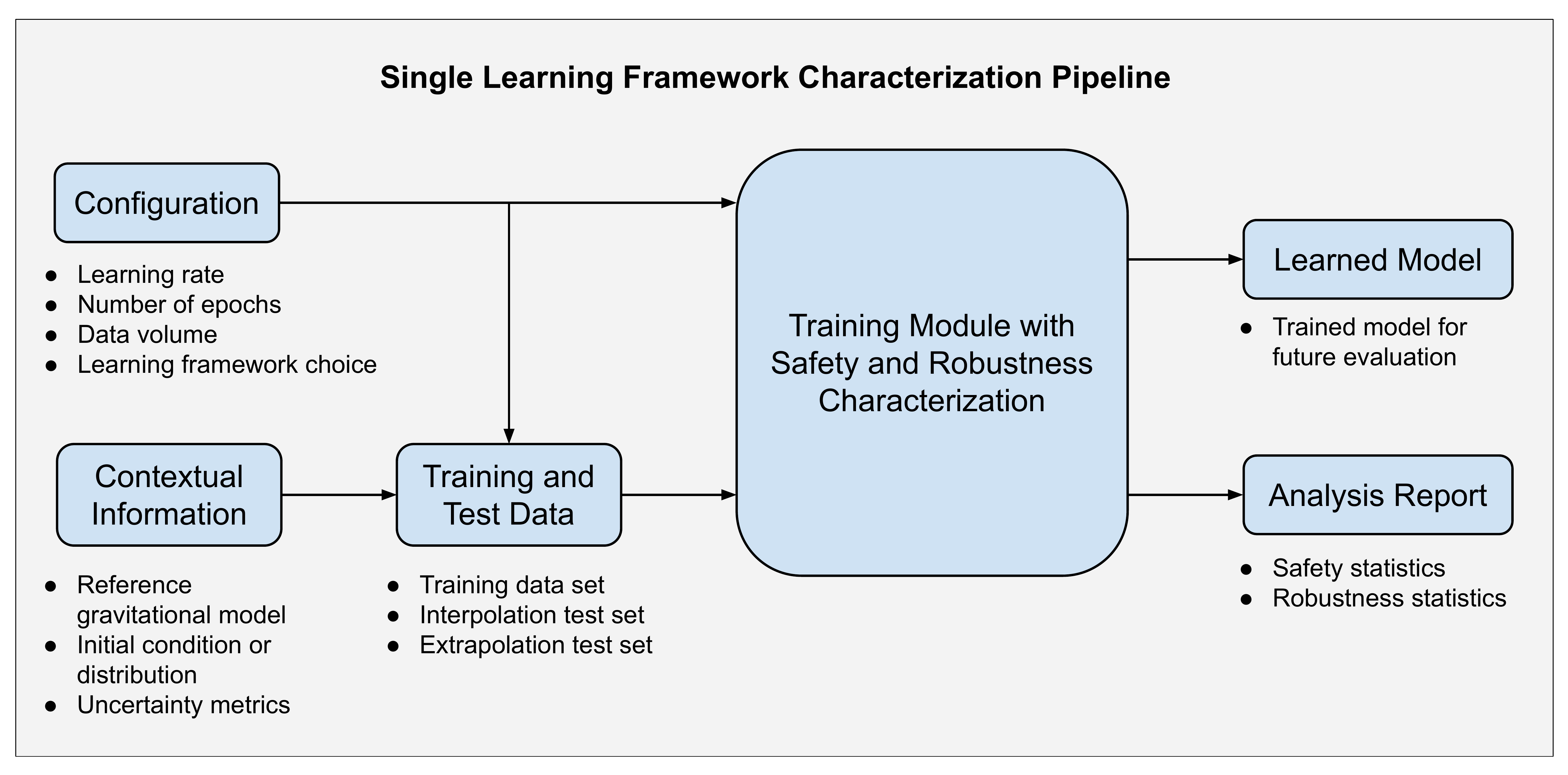}
    \caption{Pipeline architecture to characterize a single learning framework. From the inputs (Configuration and Contextual Information), the pipeline handles the data generation, learning, and characterization. The pipeline outputs a learned model that can then be used for further trajectory planning or control. It also outputs the analysis report that contextualizes the learned gravity model performance.  }
    \label{fig:singlepipe}
\end{figure}

Our single learning framework characterization pipeline has six modules as enumerated below and illustrated in Fig.~\ref{fig:singlepipe}. 
\begin{enumerate}
    \item Configuration
    \item Contextual information
    \item Training and test data
    \item Training module with safety and robustness characterization
    \item Learned model
    \item Analysis report
\end{enumerate}

The configuration and contextual information are the primary inputs. The learned model and analysis report are the primary outputs. First, the configuration is a bundle of top-level parameters. Most notably, the configuration includes the learning rate, maximum number of training epochs, data volume, and learning framework choice. Some parameters exclusively affect training (e.g., the learning rate and the maximum number of training epochs). Other parameters, like data volume and learning framework choice, affect both training and the data sets. Second, the contextual information sets the physical parameters, such as the gravitational information, initial conditions, and the uncertainty or noise to add to the trajectories. Third, we use the contextual information with the method discussed in Section~\ref{subsec:datagen} to generate the training and test data sets. In a real system, the spacecraft would use its past trajectories rather than a simulated data set. As depicted in Fig.~\ref{fig:singlepipe}, the configuration parameters may impact the training and test data sets. For example, a physics-informed neural network benefits from an estimate of the gravitational potential, in addition to the acceleration data \cite{Martin2020Applications21-393}. The data volume specification within the configuration also determines how long to propagate the trajectory during the data generation process. Equipped with learning parameters and the training data, we run training and develop the learned model (fourth module). We use the training data, interpolation test data, and extrapolation test data to generate the safety and robustness characterization report. At the end of the characterization module, we have a learned model that we can use for trajectory planning and control (fifth module). We also have a characterization report that contextualizes the accuracy of the gravity predictions from the learned model (sixth module).

To provide a more consistent metric of performance, we report the accuracy in terms of fractional error ($\epsilon$) per Eq.~(\ref{eq:frac_error}). A fractional error of $\epsilon \approx 10^{-2}$ means that the residual between the true and predicted acceleration is approximately 1\% the magnitude of the true acceleration. 
\begin{equation}
    \epsilon = \frac{||a_{\text{True}} - a_{\text{Predicted}}||}{||a_{\text{True}}||} \label{eq:frac_error}
\end{equation}

\begin{figure}
    % Insert the legend for the state data separately because I can't put a legend on a 3D plot in Matplotlib
    \includegraphics[width = 0.2\textwidth, trim={3.1cm 1.9cm 6.1cm 8.2cm}, clip]{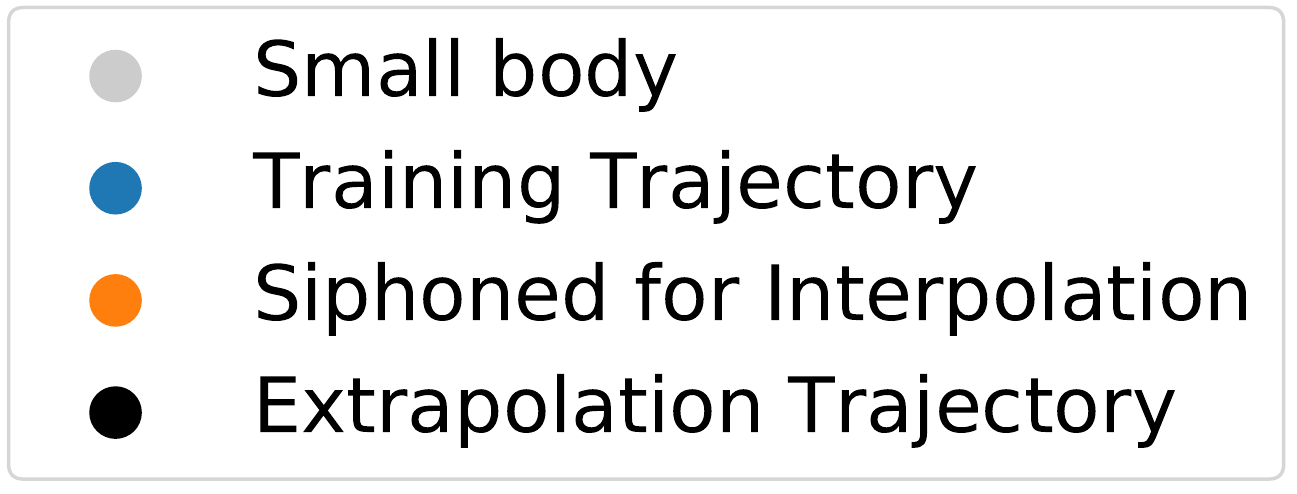}
    % \hspace*{\fill}
    
    \centering
    \begin{subfigure}{0.4 \textwidth}
         \centering
         \includegraphics[width = \textwidth, trim={4cm 0 2cm 1.5cm}, clip]{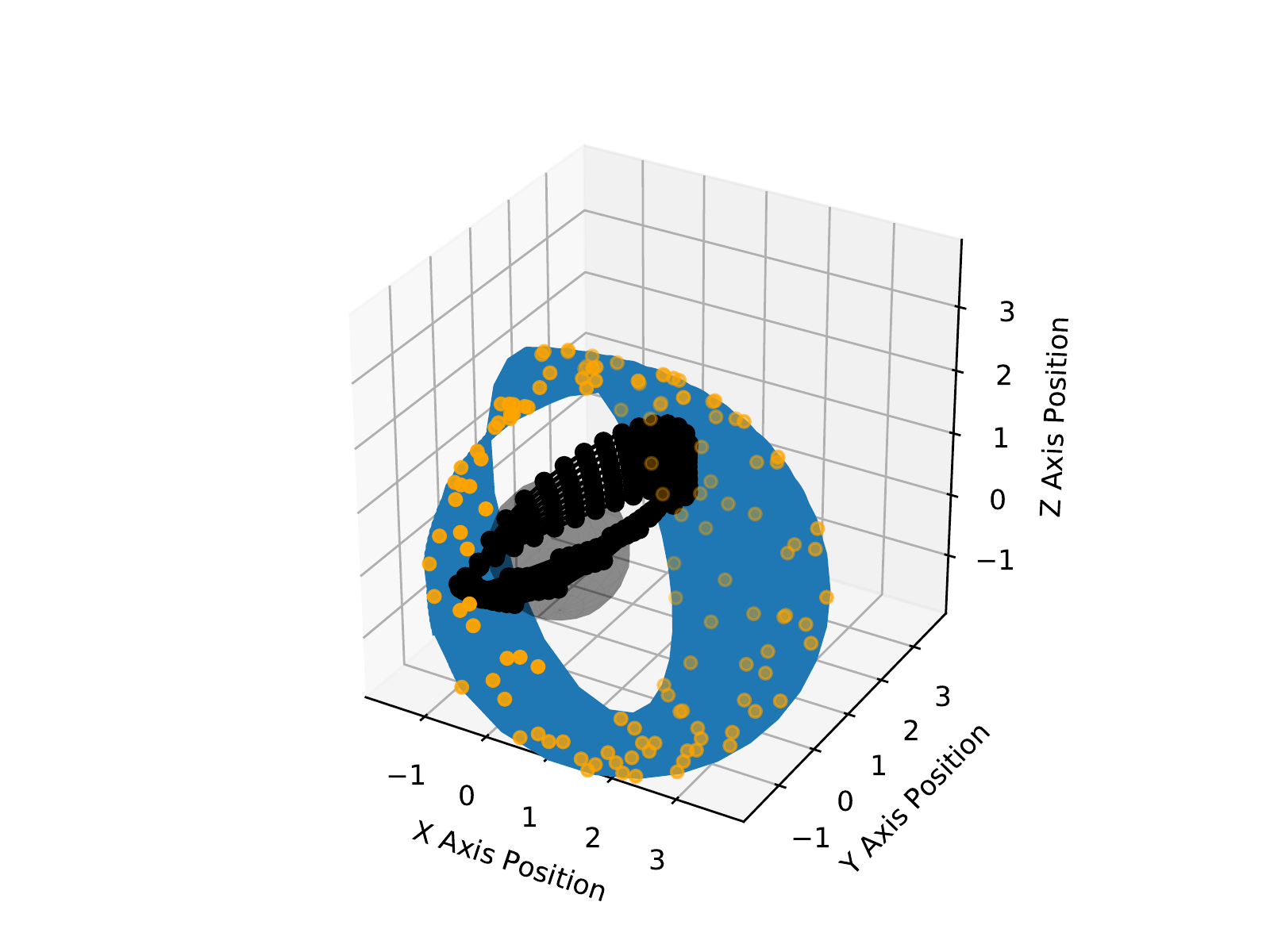}
         \caption{Train and test trajectories with normalized \\ position. Initial condition 1.}
    \end{subfigure}
    \hfill
    \begin{subfigure}{0.55 \textwidth}
         \centering
         \includegraphics[width = \textwidth]{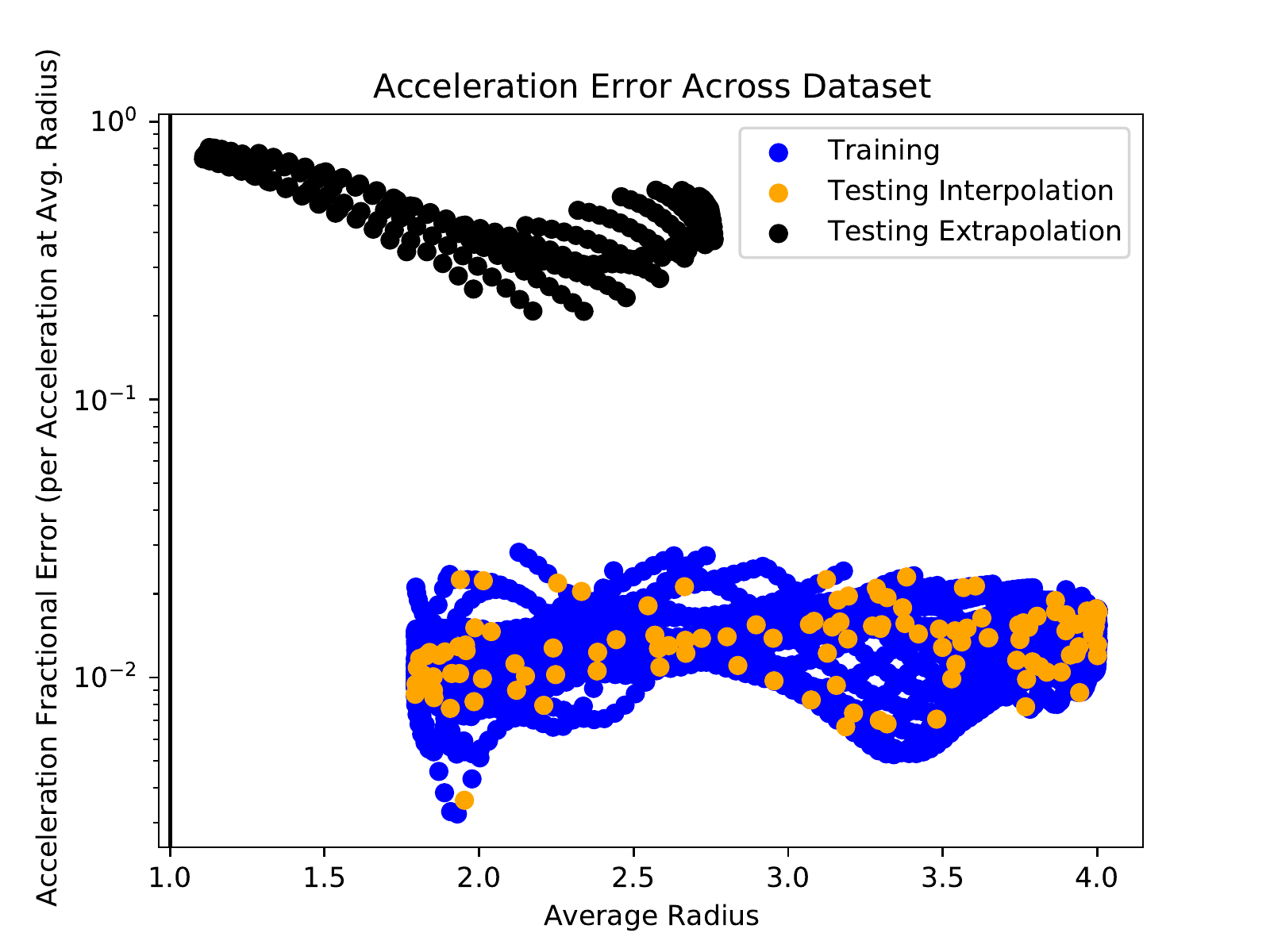}
         \caption{Fractional acceleration error for train and test trajectories. \\ Initial condition 1.}
    \end{subfigure}
    \begin{subfigure}{0.4 \textwidth}
         \centering
         \includegraphics[width = \textwidth, trim={4cm 0 2cm 1.5cm}, clip]{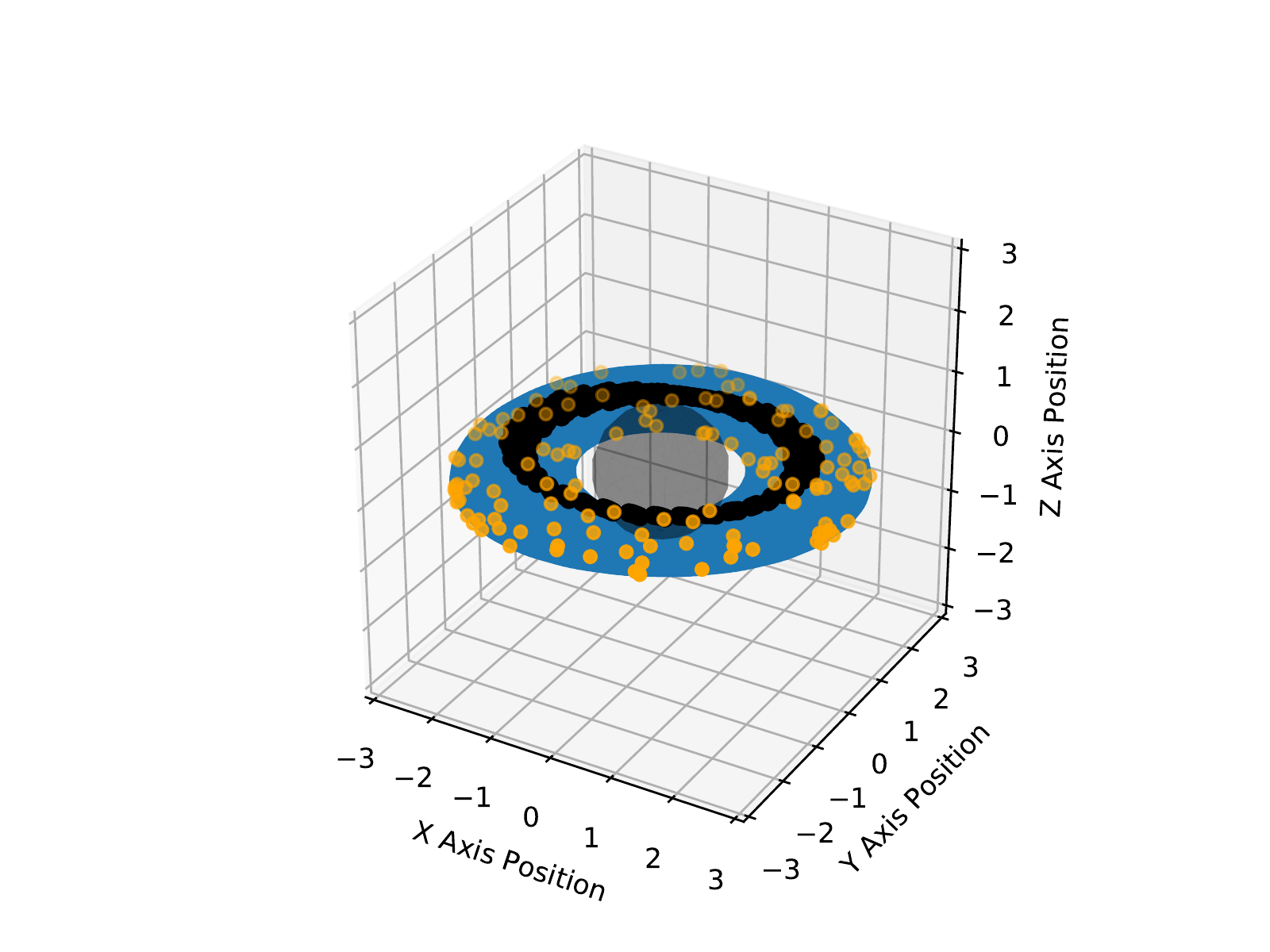}
         \caption{Train and test trajectories with normalized \\ position. Initial condition 2.}
    \end{subfigure}
    \hfill
    \begin{subfigure}{0.55 \textwidth}
         \centering
         \includegraphics[width =  \textwidth]{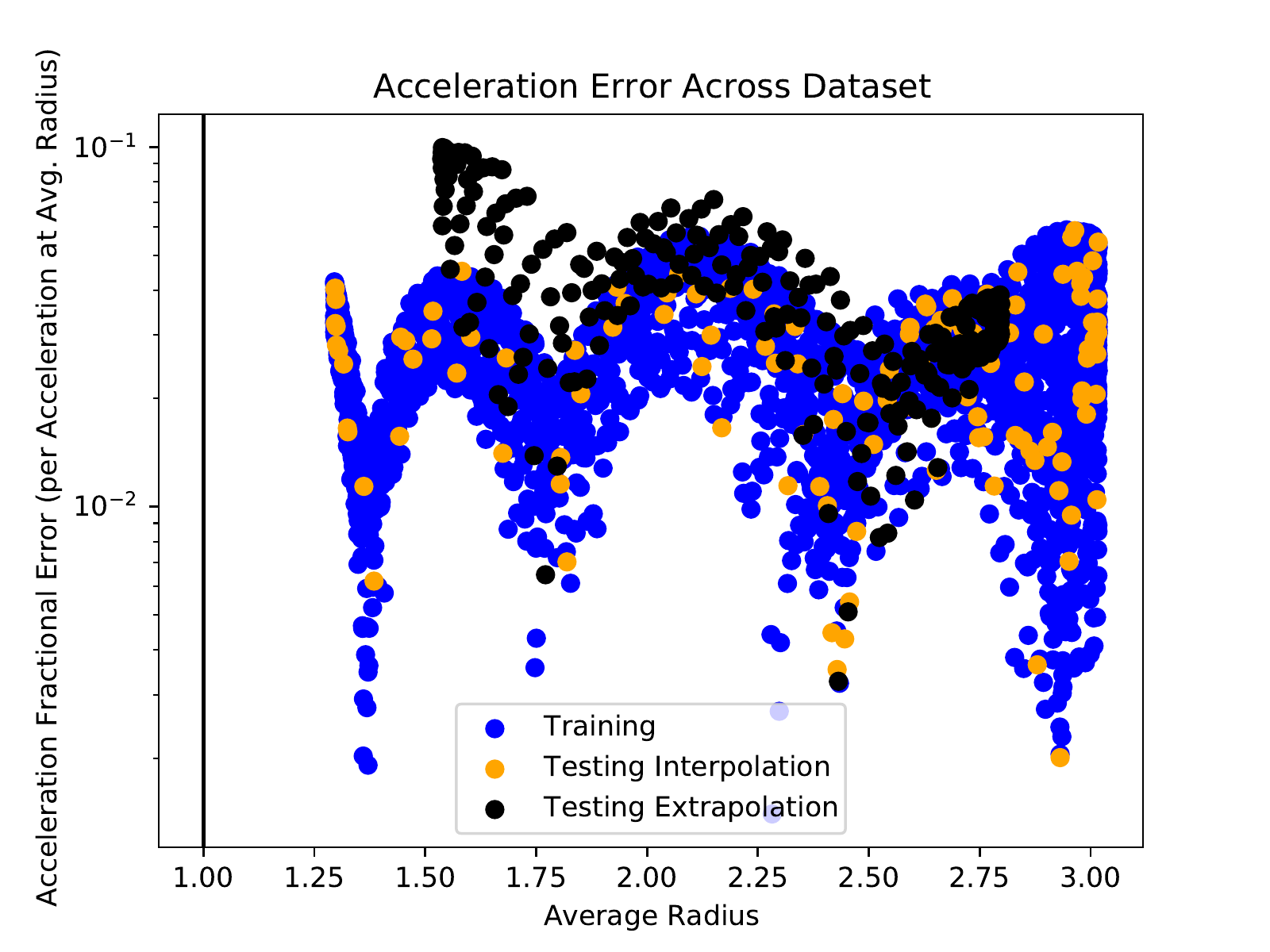}
         \caption{Fractional acceleration error for train and test trajectories. \\ Initial condition 2.}
    \end{subfigure}
    \caption{Two contrasting examples of safety and robustness characterization of a Gaussian process model with low uncertainty in position and no uncertainty in acceleration. The top row (a and b) illustrates a case where the training set and the extrapolation set do not overlap significantly in their trajectories (a). The learned model does not correctly reason about the test trajectory. Hence, there is a large gap in acceleration prediction error between the training and extrapolation testing datasets. We characterize this model as not robust, given this test trajectory. The bottom row (c and d) illustrates a case where the training set and the extrapolation set have noticeable overlap in their trajectories (c). Overall, the gap in the acceleration prediction error is minimal between the training and extrapolation testing datasets. We characterize this model as more robust, given this test trajectory.}
    \label{fig:single-char-gp}
\end{figure}

The interpolation test set consists of points siphoned from the training dataset. The interpolation test set can be viewed as an evaluation of performance within the training domain. For a safer model, the fractional error of the interpolation test set should be closer to the fractional error of the training data set. The extrapolation test set consists of a completely separate trajectory but generated within the same orbital parameter ranges as the training data set (see section~\ref{subsec:datagen}). For a more robust model, the fraction error of the extrapolation test set should be closer to the fractional error of the training data set. In Fig.~\ref{fig:single-char-gp}, we illustrate the spectrum of results for single instance characterization for a Gaussian process model under low uncertainty. The left column of Fig.~\ref{fig:single-char-gp} is the input dataset corresponding with ``Training and Test Data'' in Fig.~\ref{fig:singlepipe}. The right column of Fig.~\ref{fig:single-char-gp} is the output report corresponding with ``Analysis Report'' in Fig.~\ref{fig:singlepipe}. In both cases, we see safe predictions. The fractional error of the interpolation test set is completely overlapping with the training set. However, the robustness is more spread out. In the first case of Fig.~\ref{fig:single-char-gp}, the gap between the fractional error of the training set and the extrapolation test set is large. In such a case, a trajectory maneuver from the training set (i.e., the current and past trajectories) to the illustrated test trajectory could be catastrophic. The learned model fails to robustly generalize to the new test trajectory. On the other hand, the second case of Fig.~\ref{fig:single-char-gp} exhibits a much smaller gap between the training set and the extrapolation test set. With this learned Gaussian process model, the spacecraft can robustly transition from the training set to the test set. 

\begin{figure}
    % Insert the legend for the state data separately because I can't put a legend on a 3D plot in Matplotlib
    \includegraphics[width = 0.2\textwidth, trim={3.1cm 1.9cm 6.1cm 8.2cm}, clip]{FullPaperFigures/Single/Legend/TrajectoryLegend2.pdf}
    % \hspace*{\fill}
    
    \centering
    \begin{subfigure}{0.4 \textwidth}
         \centering
         \includegraphics[width = \textwidth, trim={4cm 0 2cm 1.5cm}, clip]{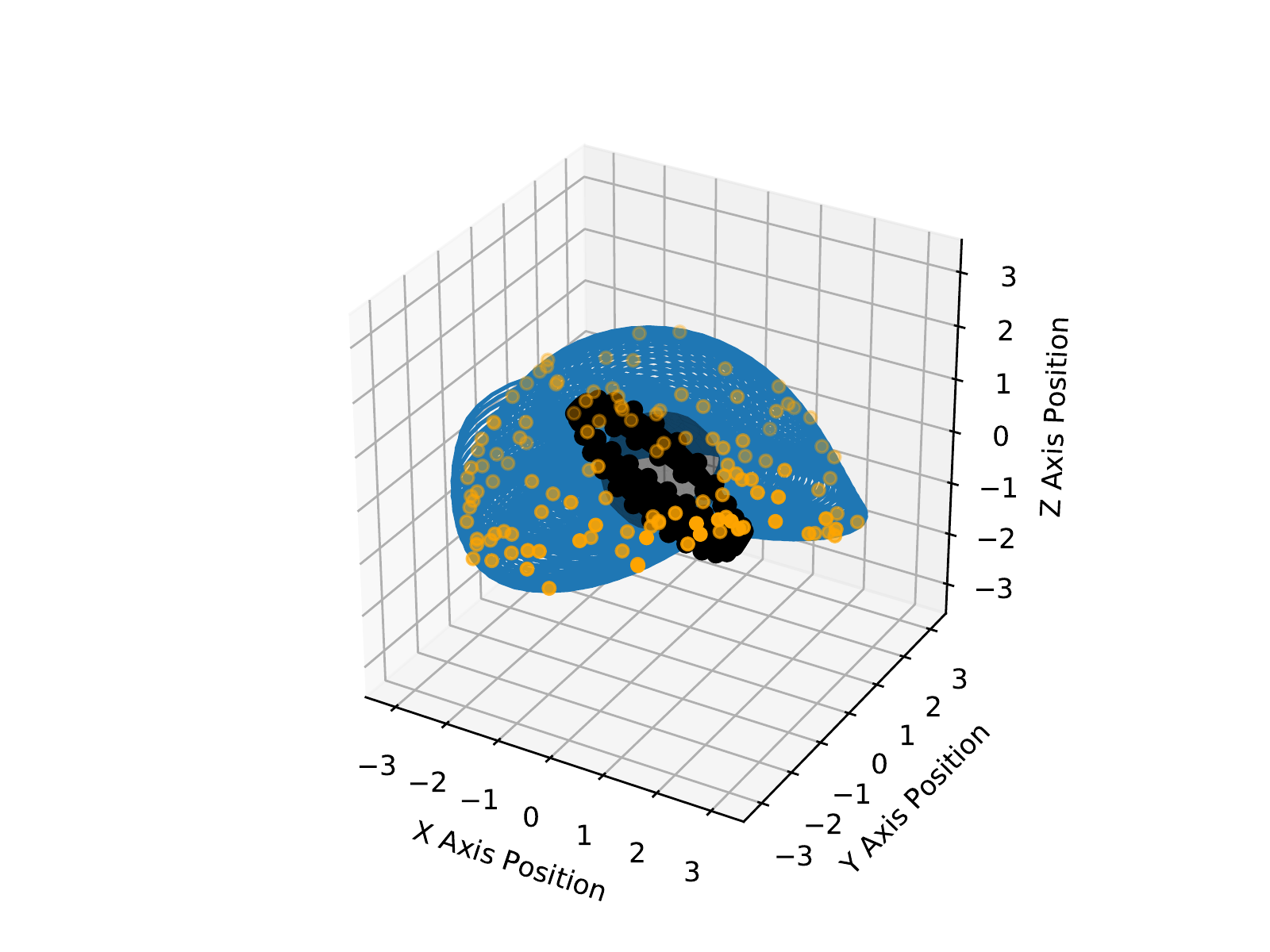}
         \caption{Train and test trajectories with normalized \\ position. Initial condition 1.}
    \end{subfigure}
    \hfill
    \begin{subfigure}{0.55 \textwidth}
         \centering
         \includegraphics[width = \textwidth]{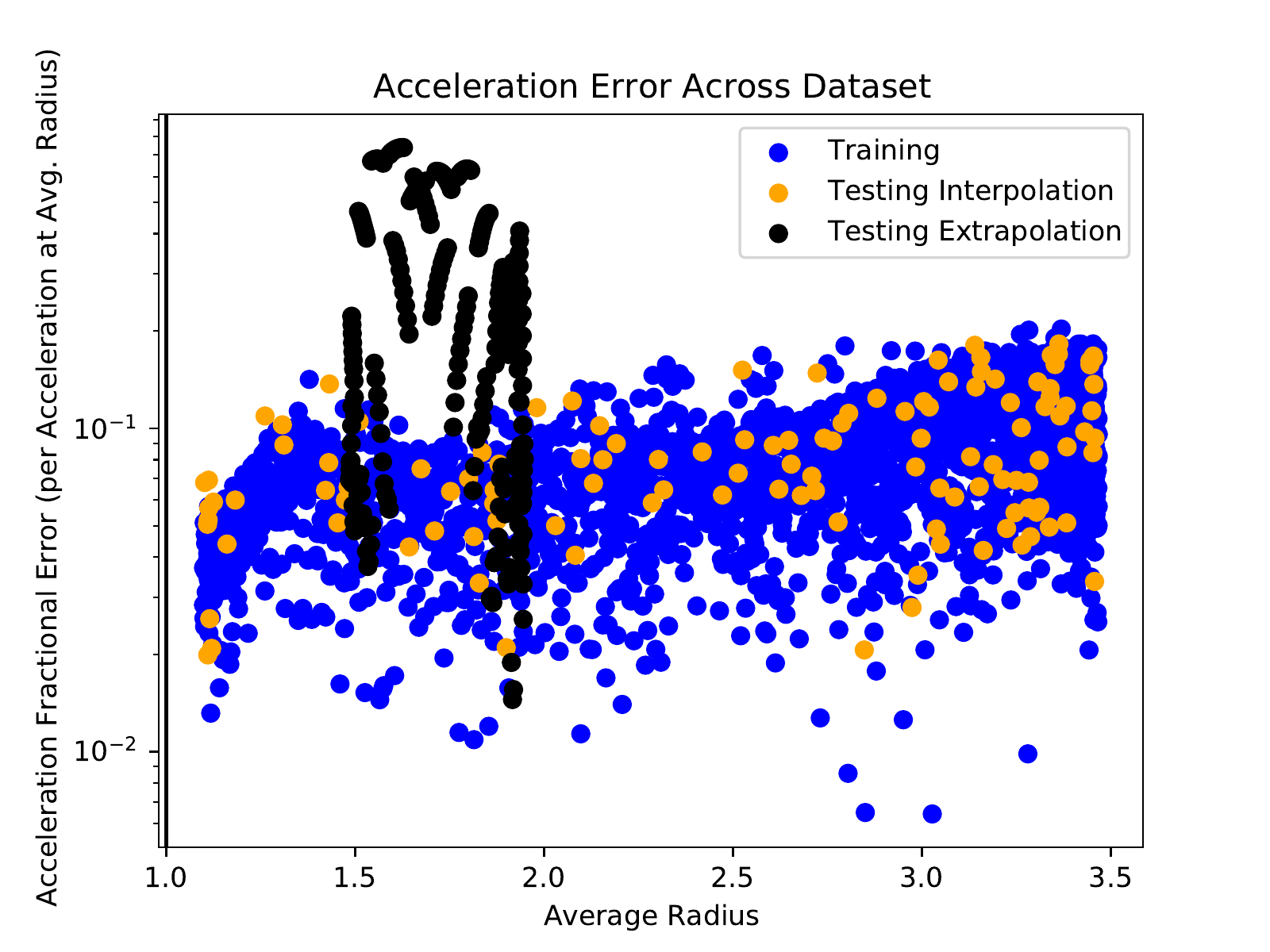}
         \caption{Fractional acceleration error for train and test trajectories.\\ Initial condition 1.}
    \end{subfigure}
    \begin{subfigure}{0.4 \textwidth}
         \centering
         \includegraphics[width = \textwidth, trim={4cm 0 2cm 1.5cm}, clip]{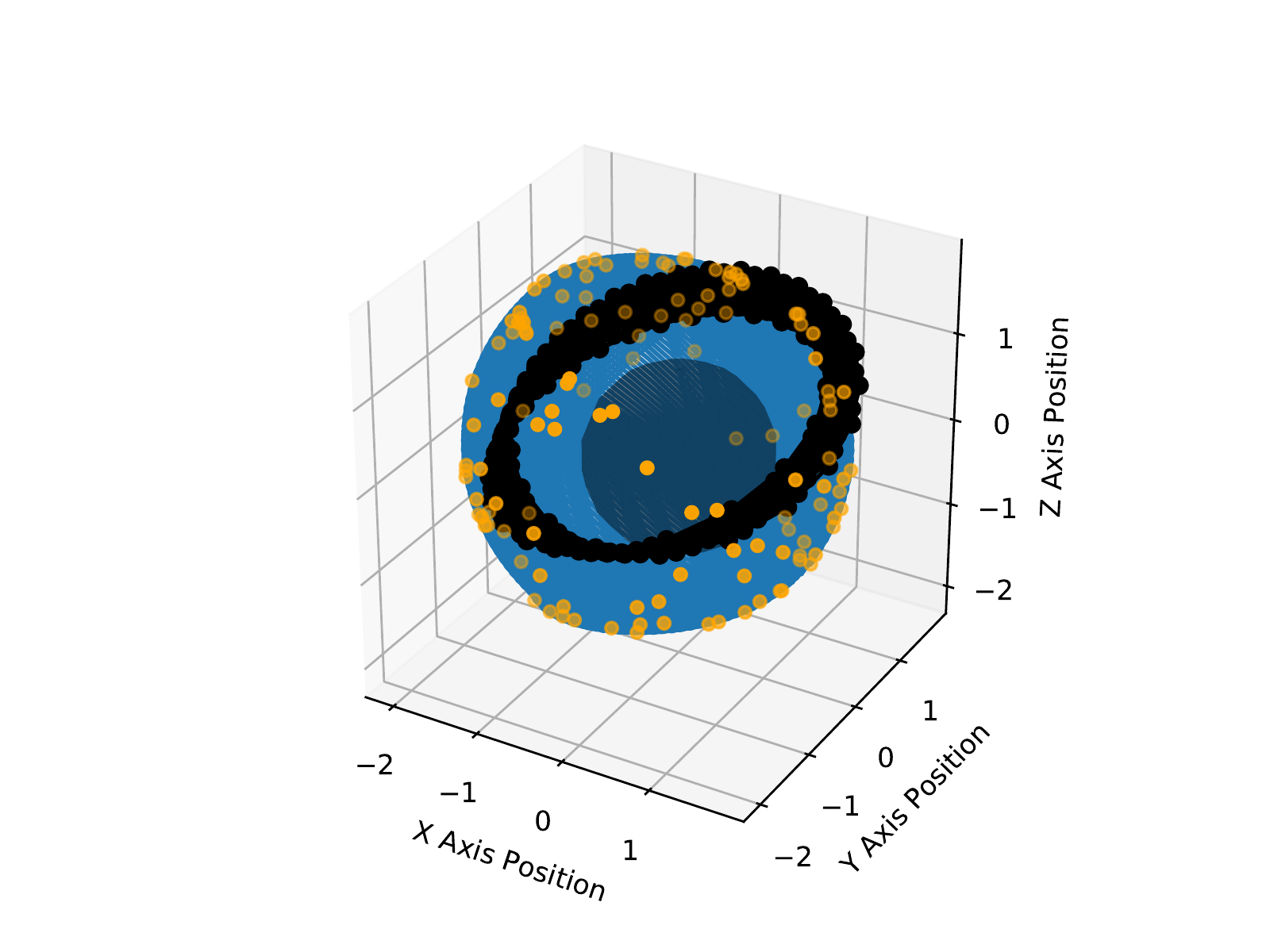}
         \caption{Train and test trajectories with normalized \\ position. Initial condition 2.}
    \end{subfigure}
    \hfill
    \begin{subfigure}{0.55 \textwidth}
         \centering
         \includegraphics[width =  \textwidth]{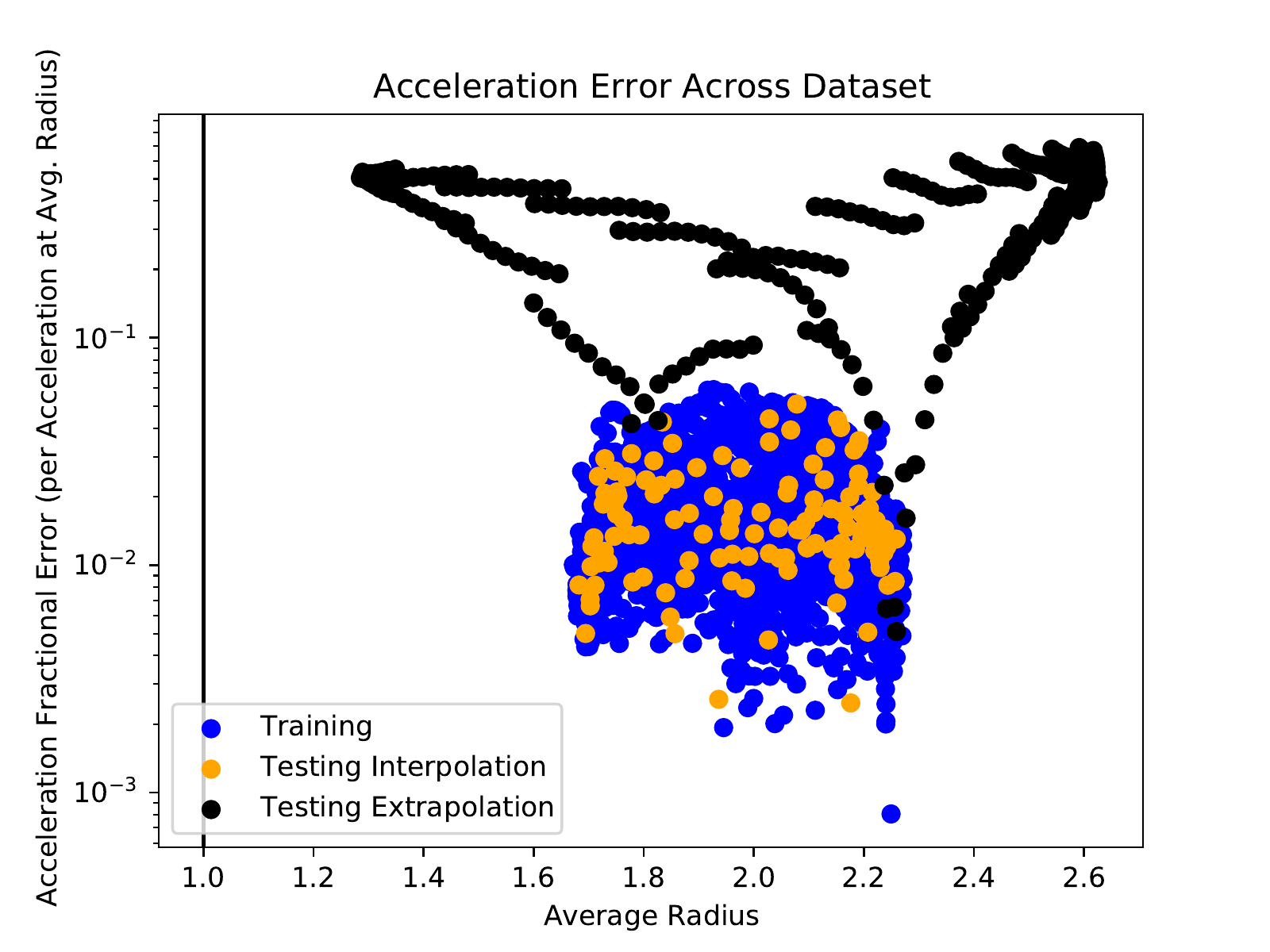}
         \caption{Fractional acceleration error for train and test trajectories. Initial condition 2.}
    \end{subfigure}
    \caption{Two contrasting examples of safety and robustness characterization of a Neural Network model with low uncertainty in position and no uncertainty in acceleration. Note that the training and testing datasets are different than the ones presented in Fig. \ref{fig:single-char-gp}. The top row (a and b) illustrates a case where the training set and the extrapolation set overlap over a section in their trajectories (a). The learned model correctly reasons about the test trajectory when the trajectory passes into the training domain. But, the gap in acceleration prediction error between the training and extrapolation testing datasets increases outside the training domain. We characterize this model as moderately robust, given this test trajectory. The bottom row (c and d) illustrates a case where the training set and the extrapolation set are nearby but barely overlap in their trajectories (c). The gap in the acceleration prediction error is minimal between the training and extrapolation testing datasets in some patches. But, overall, the gap is large outside the training domain (radii below 1.6 and greater than 2.3). We characterize this model as less robust, given this test trajectory.}
    \label{fig:single-char-nn}
\end{figure}

Similar results hold for the neural network models at low uncertainty. In Fig.~\ref{fig:single-char-nn}, we illustrate the single instance characterization for a neural network under low uncertainty. We selected two cases where the training set and the extrapolation set partially overlap to demonstrate the performance as the extrapolation trajectory enters and leaves the training domain. As with the Gaussian process models, we see safe predictions where the error in the interpolation set is comparable to the training set. The models are fairly robust for sections of the trajectory that pass near or into the training domain. However, the prediction accuracy drops quickly outside of the training domain. The second case (Fig.~\ref{fig:single-char-nn}c and d) exhibits a noticeable accuracy drop outside of the training domain. In this case, the extrapolation test set is quite close to the training set, yet the predictions are not as accurate as the predictions in the training set. We would characterize the first case (subfigures a and b) as more robust than the second case (subfigures c and d) for the selected test trajectories. 

In both of the Gaussian process and neural network cases, we considered a single test trajectory for simpler visualization. Depending on the selection of training and test trajectories, we can arrive at different robustness characterization metrics. Using many extrapolation test trajectories would better characterize a learning-based technique's robustness. Alternatively, the extrapolation test trajectory might be user-selected based on the needs of the mission. Such orbits include those necessary for landing, target observation, or Earth ground station communication.

% Testing for the legend for the states above
% \begin{figure}
%     \centering
% %     {%
% % \setlength{\fboxsep}{0pt}%
% % \setlength{\fboxrule}{1pt}%
% % \fbox{\includegraphics[width = \textwidth, trim={3.1cm 1.9cm 6.1cm 8.2cm}, clip]{FullPaperFigures/Single/Legend/TrajectoryLegend2.pdf}}%
% % }%
%     \includegraphics[width = \textwidth, trim={3.1cm 1.9cm 6.1cm 8.2cm}, clip]{FullPaperFigures/Single/Legend/TrajectoryLegend2.pdf}
%     \caption{Caption}
%     \label{fig:my_label}
% \end{figure}

\newpage
\subsection{Framework Characterization}

\begin{figure}
    \centering
    \includegraphics[width=0.95\textwidth]{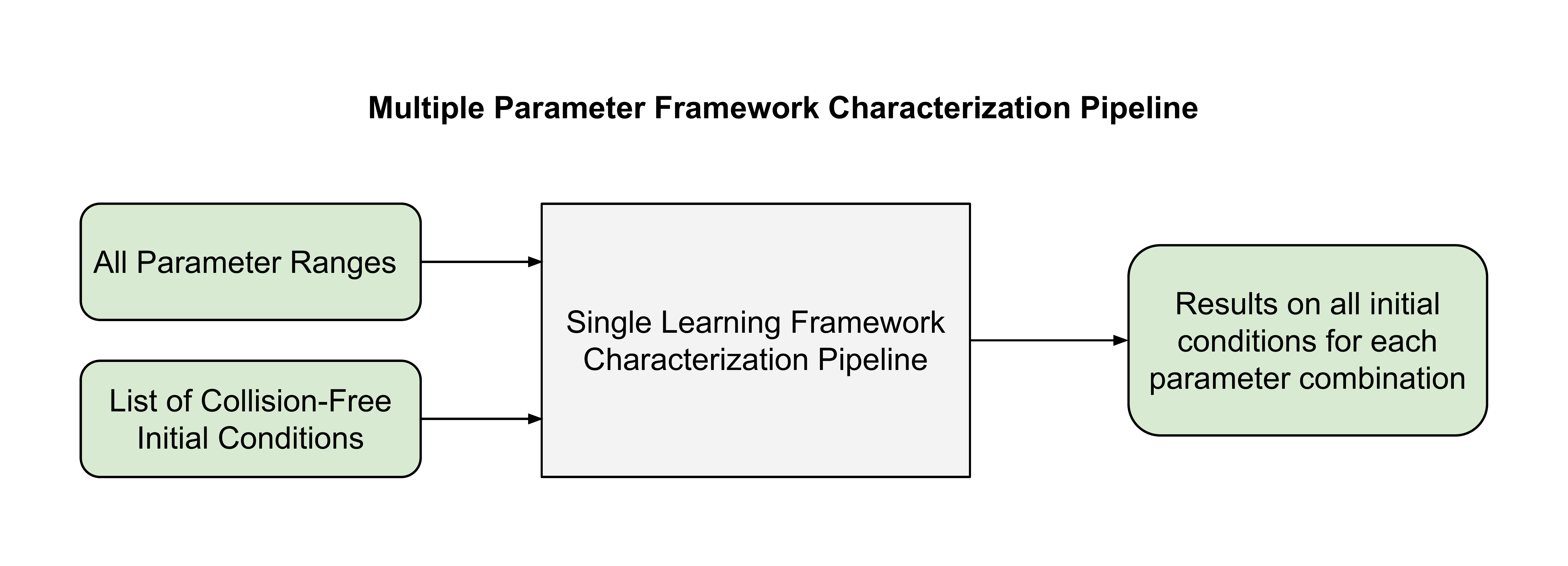}
    \caption{Pipeline architecture to characterize a collection of frameworks over multiple parameters. The central block ``Single Learning Framework Characterization Pipeline'' is the pipeline demonstrated in Fig.~\ref{fig:singlepipe}. The user specifies the full parameter range and spacing for each parameter (e.g., uncertainty magnitude of (0.1, 0.3, 0.5) and (Gaussian process, neural network) frameworks. The pipeline then passes each instance of the parameters to the configuration or contextual information of the ``Single Learning Framework Characterization Pipeline'' (see Fig.~\ref{fig:singlepipe}). We separate the list of collision-free initial conditions to ensure repeatability for each parameter instance. The pipeline outputs the aggregated results of each parameter instance over the range of collision-free initial conditions.}
    \label{fig:varyingpipe}
\end{figure}

Our full framework characterization pipeline has four modules (Fig.~\ref{fig:varyingpipe}).
\begin{enumerate}
    \item Parameter ranges
    \item List of collision-free initial conditions
    \item The single learning framework characterization pipeline
    \item The pipeline results
\end{enumerate}

A separated input architecture ensures a more repeatable comparison between safety and robustness characterization. Specifically, we evaluate every realization of the parameters on the same initial conditions. As such, we store the initial conditions separately. The parameter ranges specify how to vary all the remaining parameters. Each realization in the parameter range represents a complete configuration and contextual information instance, except for the initial condition. We then input the initial conditions and this realization into the single learning framework characterization pipeline. Selecting large parameter ranges and many initial conditions quickly increases the total number of runs of the single learning framework characterization pipeline. For example, suppose we select two learning frameworks, ten state uncertainty values, and 100 initial conditions. Then, we have to run 2000 runs of the single learning framework characterization pipeline. For large data volumes, a mere 2000 runs can take significant amounts of time. Large-scale characterization is hence best suited for a computing server, especially at large data volumes.

\begin{figure}
    \centering
    % \begin{subfigure}{0.52 \textwidth}
    %      \centering
    %      \includegraphics[width = \textwidth]{FullPaperFigures/Multiple/DirectCompareModerateData/TrainingFrameworkCompare.pdf}
    %      \caption{Training Error Comparison}
    % \end{subfigure}
    % \hfill
    \begin{subfigure}{0.49 \textwidth}
         \centering
         \includegraphics[width = \textwidth]{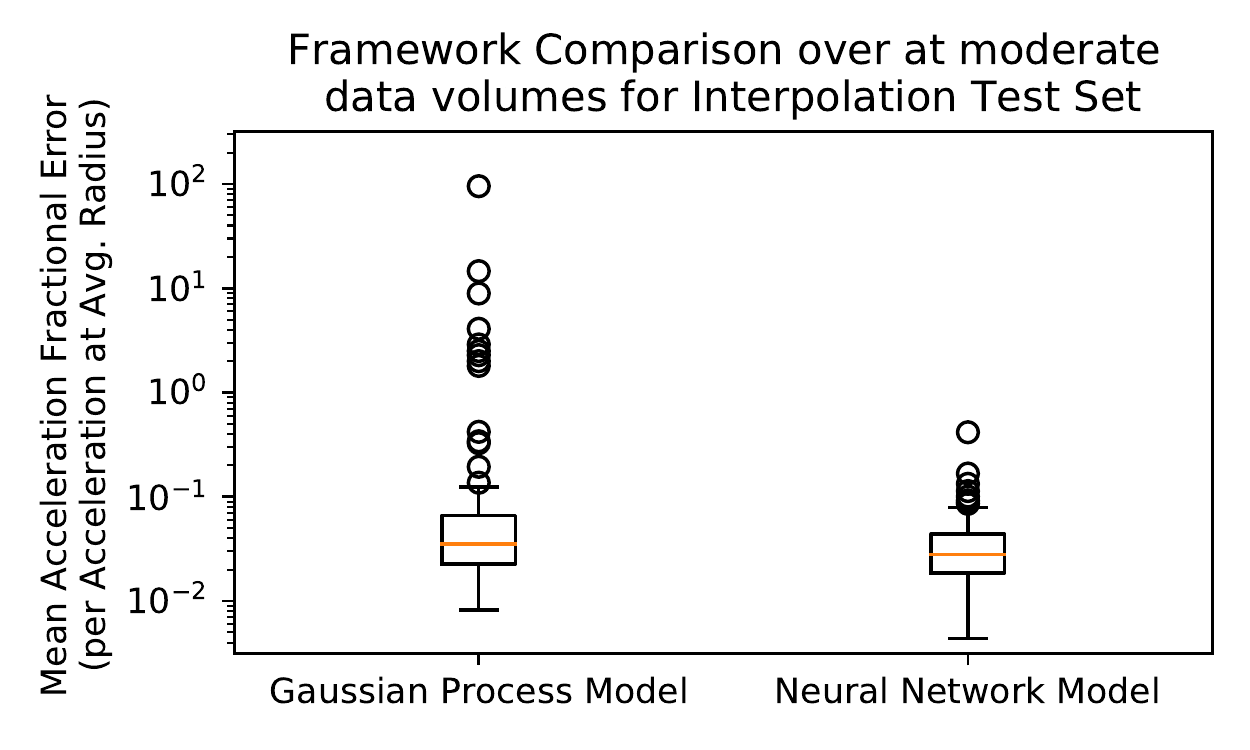}
         \caption{Interpolation Error Comparison (Safety)}
    \end{subfigure}
    \begin{subfigure}{0.49 \textwidth}
         \centering
         \includegraphics[width = \textwidth]{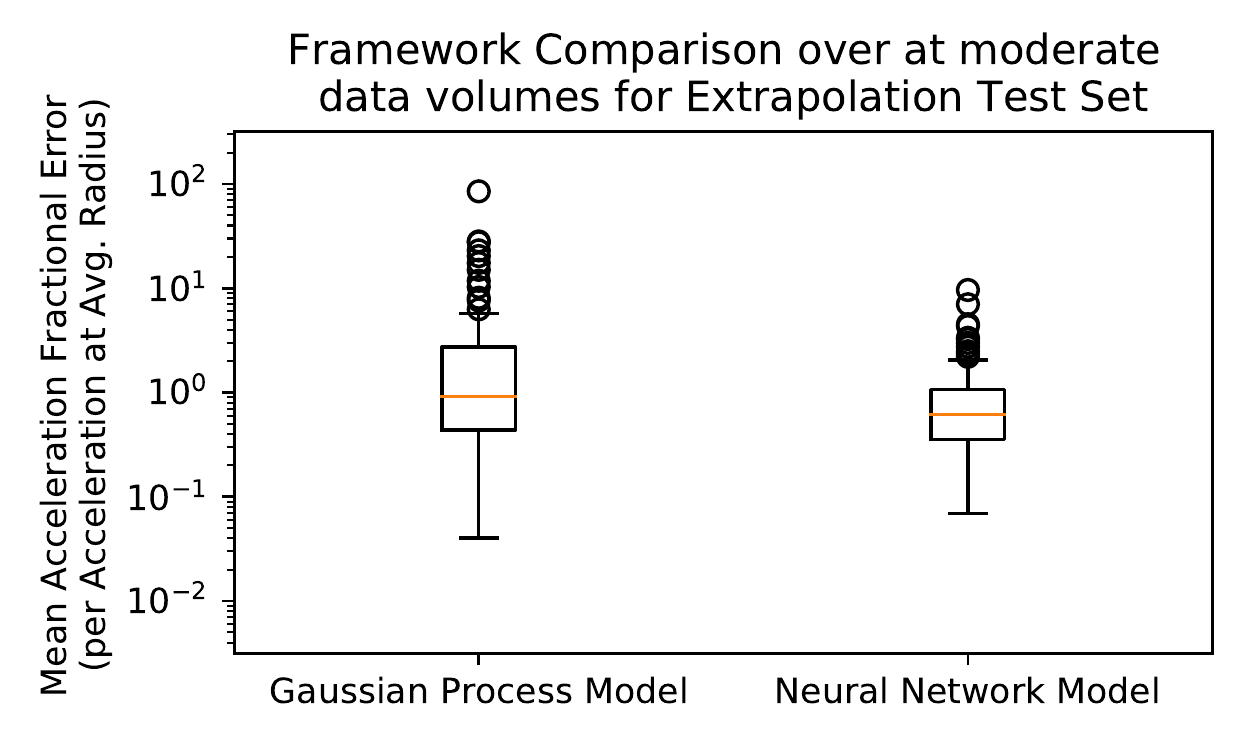}
         \caption{Extrapolation Error Comparison (Robustness)}
    \end{subfigure}
    \caption{Head-to-head comparison of the Gaussian process and neural network models in the case of moderate data volumes and minimal error. We use 100 collision-free trajectories in training and testing. Overall the Gaussian process and neural network models show comparable performance over the safety (a) and robustness (b) characterization. However, numerical instabilities in the Gaussian process model result in high error outliers.}
    \label{fig:multi-box-compare}
\end{figure}

First, we tested the multiple-parameter framework characterization pipeline with 100 collision-free initial conditions for a single parameter range at moderate data volumes. For each training set, we propagated the true orbit from one initial condition for 100 instantaneous, normalized Keplerian periods (with respect to the initial condition). We sample the orbit 25 times per period. We siphoned 5\% of this orbit for interpolation testing. We propagated the true orbit from another initial condition for 10 instantaneous, normalized Keplerian periods for the extrapolation testing set. We again sampled the orbit 25 times per period. Overall, we observed comparable performance between the Gaussian process and neural network models under these parameter conditions (Fig.~\ref{fig:multi-box-compare}). As noted in \cite{Neamati2021NewFieldsFixed}, the Gaussian process can hit numerical instabilities during training. We can attribute the outliers in the Gaussian process model between $10^{-0.5}$ and $10^2$ to this numerical instability. As noted in our example of the single learning framework characterization pipeline (Fig.~\ref{fig:single-char-gp}), there is often only a little bias between the interpolation test set and the training set performance. However, there is often a noticeable gap between the extrapolation test set performance and both the training set and interpolation test set performance. This gap is quantifiable between Fig.~\ref{fig:multi-box-compare}a and Fig.~\ref{fig:multi-box-compare}b. On average, the extrapolation test set error is one to two orders of magnitude larger than the interpolation test set error. In other words, the safety metric is generally one to two orders of magnitude better than the robustness metric.

\begin{figure}
    \centering
    \includegraphics[width=0.9\textwidth]{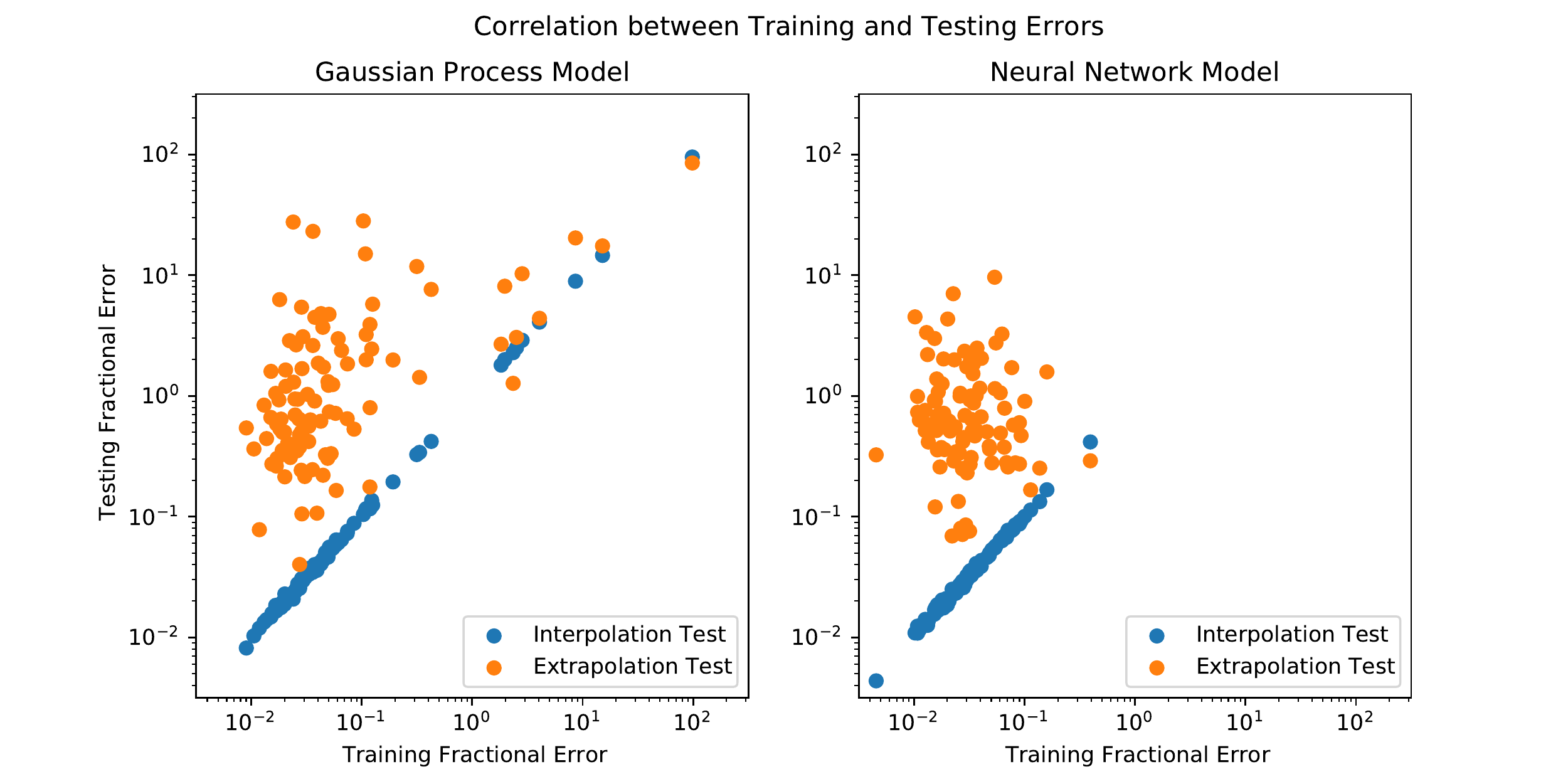}
    \caption{Assessment of correlation between the errors in training and the errors in testing. The errors in interpolation testing are very highly correlated with the errors in training (blue points) for both the Gaussian process model and the neural network model. With this correlation, we can estimate the safety characterization from the training error. In contrast, the errors in extrapolation testing are not well correlated with the errors in training (orange points) for either the Gaussian process model or the neural network model. It would be difficult to estimate the robustness characterization from the training error alone. }
    \label{fig:multi-error-correlation}
\end{figure}

We can further diagnose the correlation between the interpolation test set and the training set in comparison to the correlation between the extrapolation test set performance and the training set (Fig.~\ref{fig:multi-error-correlation}). The interpolation test set and training set errors are highly correlated for both the Gaussian process and neural network models. For the Gaussian process model, we obtained a linear fit in the logarithmic representation (as shown in Fig.~\ref{fig:multi-error-correlation}). The linear fit coefficient is 0.9995, the intercept is 0.0056, and $R^2 = 0.9992$. We repeated this linear regression for the neural network model. The linear fit coefficient is 0.9865, intercept is -0.0169, and $R^2 = 0.9954$. Not only are the coefficients of determination ($R^2$) nearly one, the slopes are also nearly one, and the intercepts are very small. So, the training error essentially matches the interpolation test set. Hence, we can reliably use the training set error to estimate the interpolation test set error. In contrast, the extrapolation test and training errors are not well correlated. Attempting a linear fit yields $R^2 = 0.3351$ for the Gaussian process model and $R^2 = 0.0076$ for the neural network. We attribute the larger coefficient of determination for the Gaussian process model to the numerical instabilities resulting in higher error. Hence, it is not possible to estimate the extrapolation test error from the training errors. So, it is not possible to directly estimate the learning model's robustness from the training performance while the spacecraft is in orbit.

% \missingfigure{FIGURES: Additional Varying Parameters Comparisons (In progress). One will be for low data volumes and increasing state uncertainty. The second will be for low data volumes and increasing acceleration uncertainty.}

Next, we tested the multiple-parameter framework characterization pipeline at low data volumes while varying state or acceleration uncertainties. We used the same 100 collision-free initial conditions as before. To use lower data volumes, we propagated the true orbit form one initial condition for 10 instantaneous, normalized Keplerian periods. As before, we sampled the orbit 25 times per period and siphoned 5\% for interpolation testing. Again, in the spirit of lower data volumes, we propagated the true orbit from another initial condition for 3 instantaneous, normalized Keplerian periods of the extrapolation testing set. We include zero-mean uncorrelated Gaussian noise to add uncertainty. The state uncertainties ($\sigma_s$) or acceleration uncertainties ($\sigma_a$) are implemented as detailed in Eq.~(\ref{eq:uncertainties}).
\begin{equation}
    a_{\text{observed}}(x_{\text{true}}) = a(x_{\text{true}} + \mathcal{N}(0, \sigma_s I)) + \mathcal{N}(0, \sigma_a I) \label{eq:uncertainties}
\end{equation}
Note that $I$ is the $3 \times 3$ identity matrix. Under uncertainty, the true state $x_{\text{true}}$ is not observed. Instead, we observe $x_{\text{observed}} = x_{\text{true}} + \mathcal{N}(0, \sigma_s I)$. Similarly, the observed acceleration $a_\text{observed}$ deviates from truth by $\mathcal{N}(0, \sigma_a I)$. During training, we provide the observed states and accelerations to the learning frameworks. The normalization technique used for state and acceleration uncertainty is meant to put that state and acceleration values at similar magnitude near unity. More details are provided in our previous work, \cite{Neamati2021NewFieldsFixed}. The magnitude of state values are generally a bit larger than unity ($\approx 1-3$). The magnitude of the acceleration values are generally larger than the state values ($\approx 1-100$).

\begin{figure}
    \centering
    \begin{subfigure}{0.49 \textwidth}
         \centering
         \includegraphics[width = \textwidth]{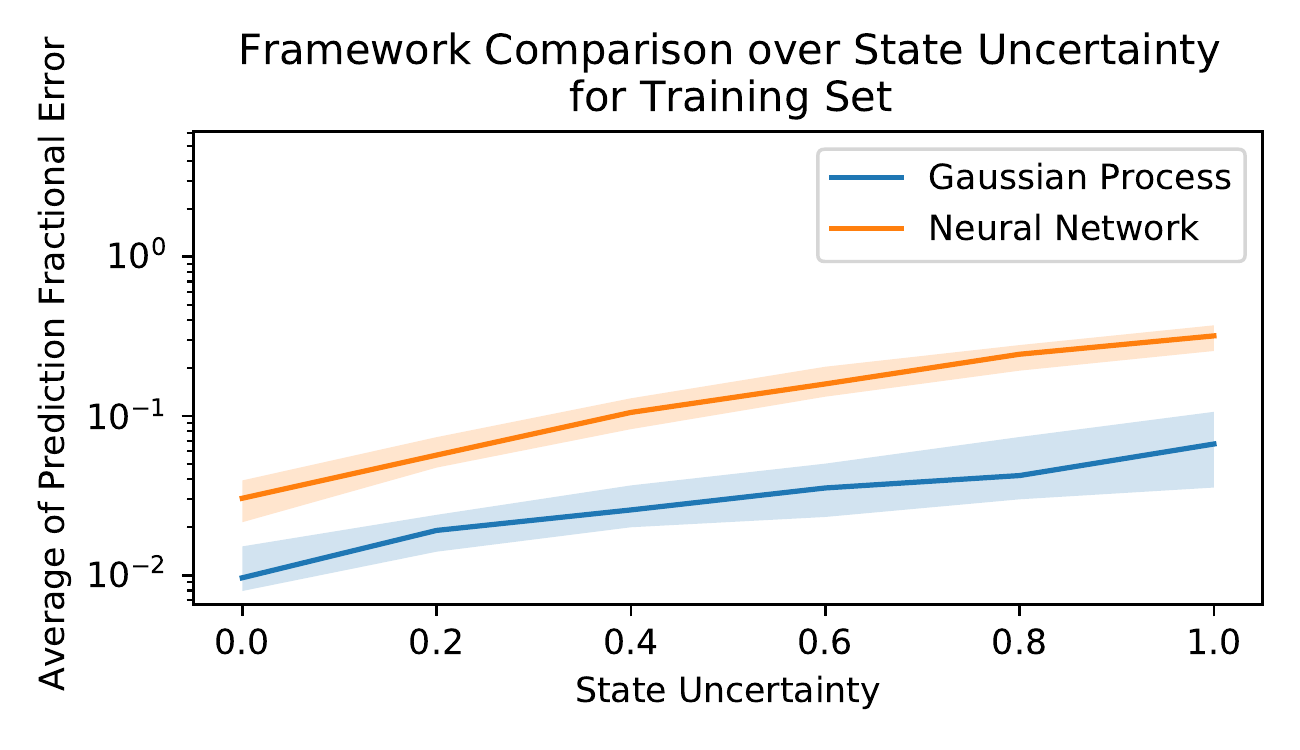}
         \caption{Training error (log scale)}
    \end{subfigure}
    \begin{subfigure}{0.49 \textwidth}
         \centering
         \includegraphics[width = \textwidth]{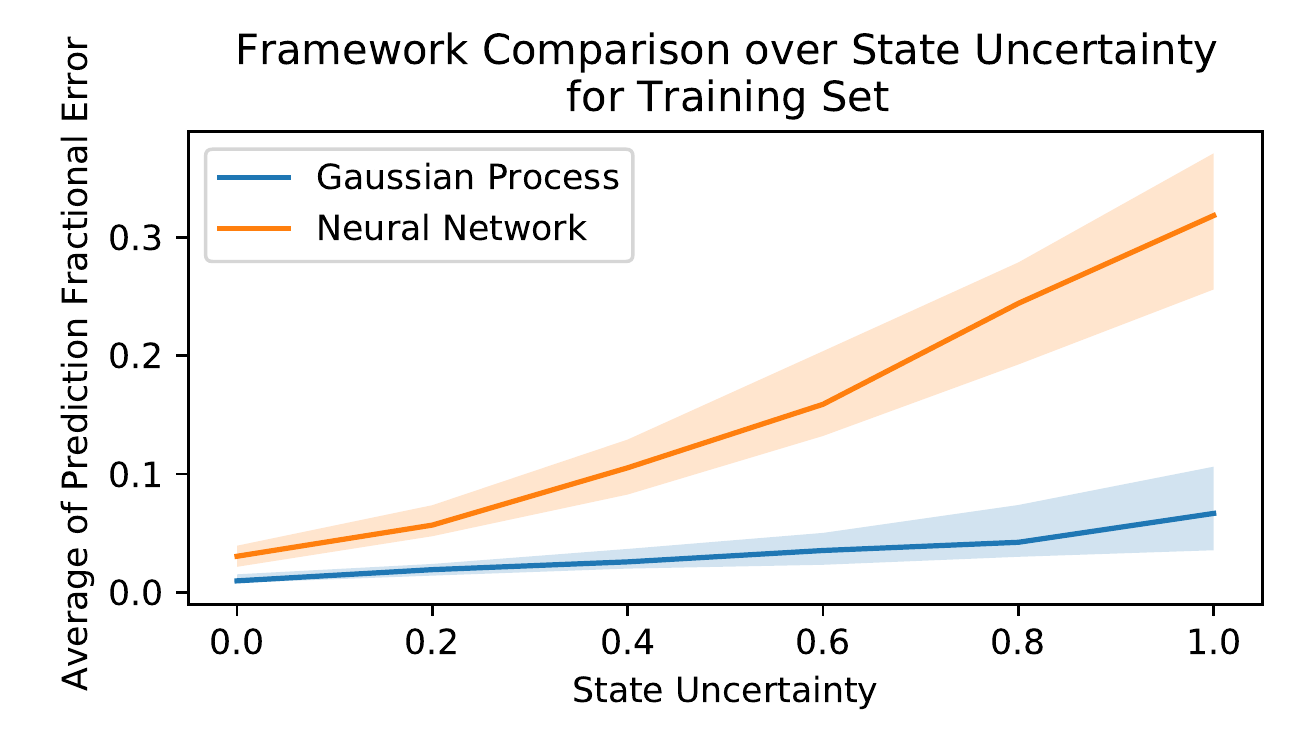}
         \caption{Training error (linear scale)}
    \end{subfigure}
    \hfill
    \begin{subfigure}{0.49 \textwidth}
         \centering
         \includegraphics[width = \textwidth]{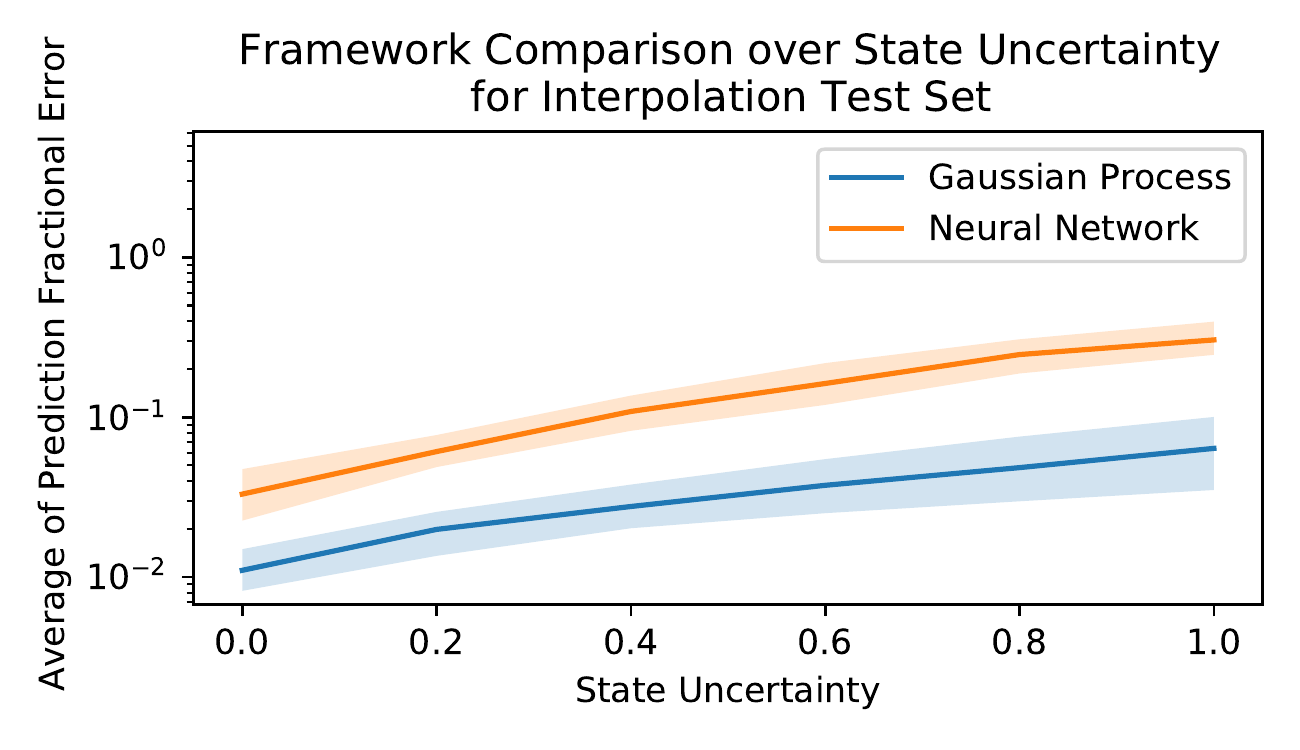}
         \caption{Interpolation test error}
    \end{subfigure}
    \begin{subfigure}{0.49 \textwidth}
         \centering
         \includegraphics[width = \textwidth]{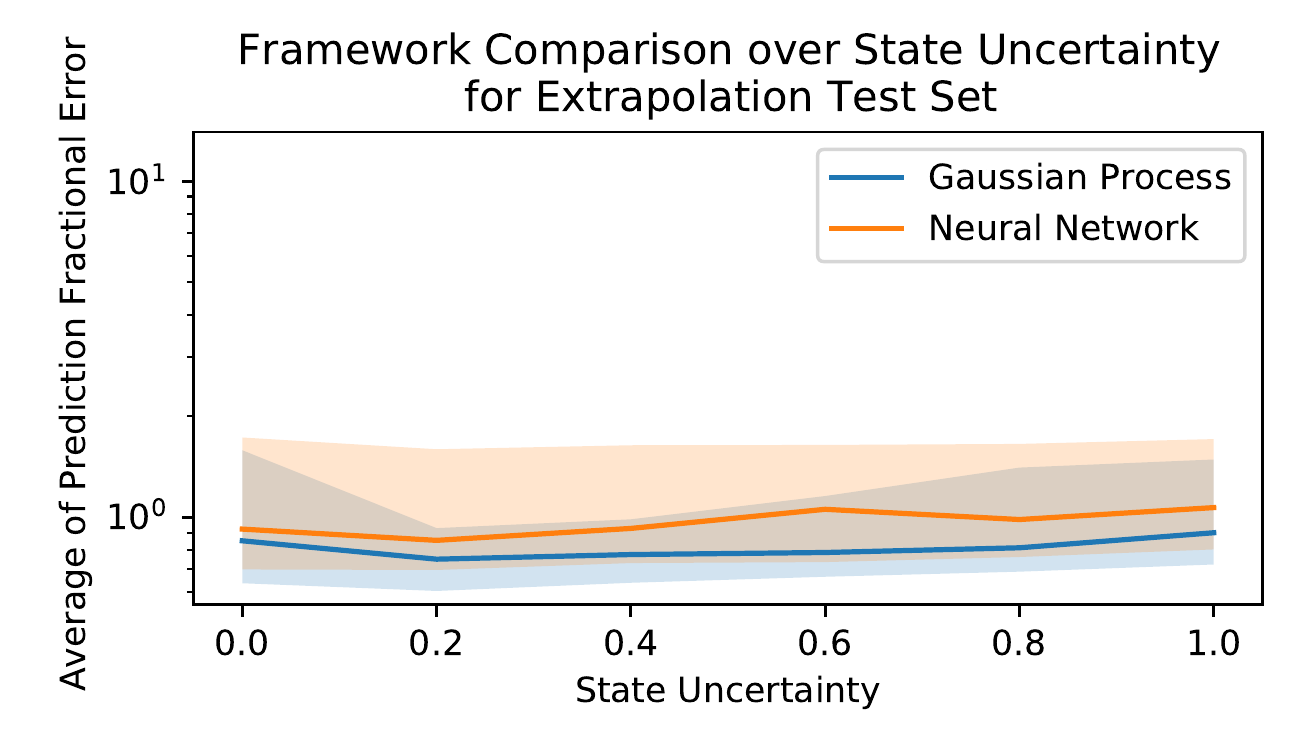}
         \caption{Extrapolation test error}
    \end{subfigure}
    \caption{Comparison of Gaussian process and neural network frameworks for a range of state uncertainties at low data volumes. The solid line is the median over 100 runs, and the translucent band is the interquartile range. We plot the training error on a log scale (a) for comparison at lower error and a linear scale (b) for comparison at higher error. The logarithmic scale bounds match Fig.~\ref{fig:accel-uncertainty}. The training error (a-b) and interpolation test error (c) nearly exponentially increase with increased state uncertainty. The extrapolation test error (d) is consistently poor throughout the range of state uncertainties. At lower data volumes (as in this figure), the Gaussian process models continuously outperform the neural network models.}
    \label{fig:state-uncertainty}
\end{figure}

In general, Gaussian Process models better handle low data volumes and higher uncertainty, as preliminarily observed in our prior work \cite{Neamati2021NewFieldsFixed}. We observe this behavior in the subsequent analysis. First, Fig.~\ref{fig:state-uncertainty} details performance as the standard deviation of the state uncertainty increases from 0 to 1. We fixed the acceleration uncertainty at no noise ($\sigma_a = 0$). The training error increases as the uncertainty increases since it is more difficult to pick out the underlying acceleration function from the noisy observed accelerations. As observed in Fig.~\ref{fig:multi-error-correlation}, the interpolation test error, which we use for safety characterization, follows the training error. As the state uncertainty increases, the safety of the learned model decreases. The extrapolation test error, which we use for robustness characterization, is consistently poor across the range of state uncertainty. With low data volumes, we cannot expect these learned models to be more robust than the learned trained on higher data volumes, as in Fig.~\ref{fig:multi-box-compare} or \ref{fig:multi-error-correlation}. The robustness does not noticeably change across state uncertainty since the system is not robust, to begin with. We conclude that the Gaussian process models are safer and more robust than the neural network models for low data volumes and higher state uncertainty.

\begin{figure}
    \centering
    \includegraphics[width=0.49 \textwidth]{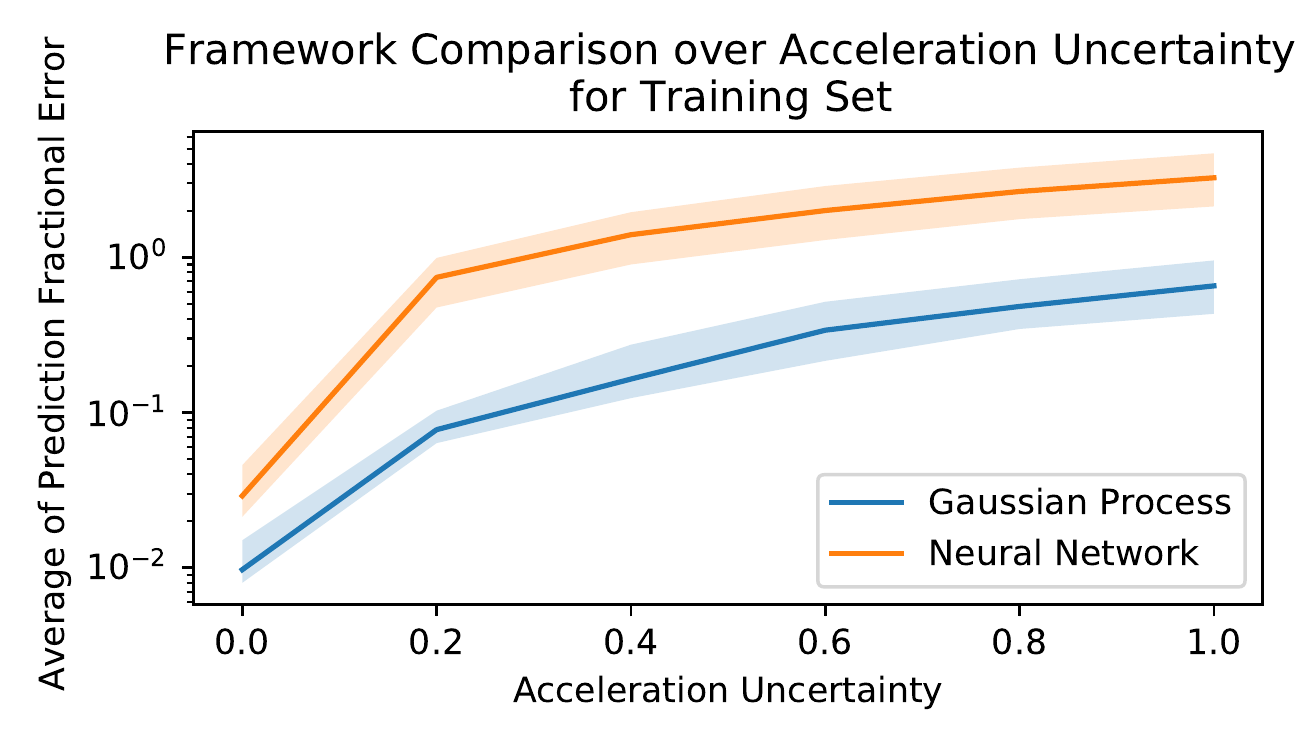}
    \includegraphics[width=0.49 \textwidth]{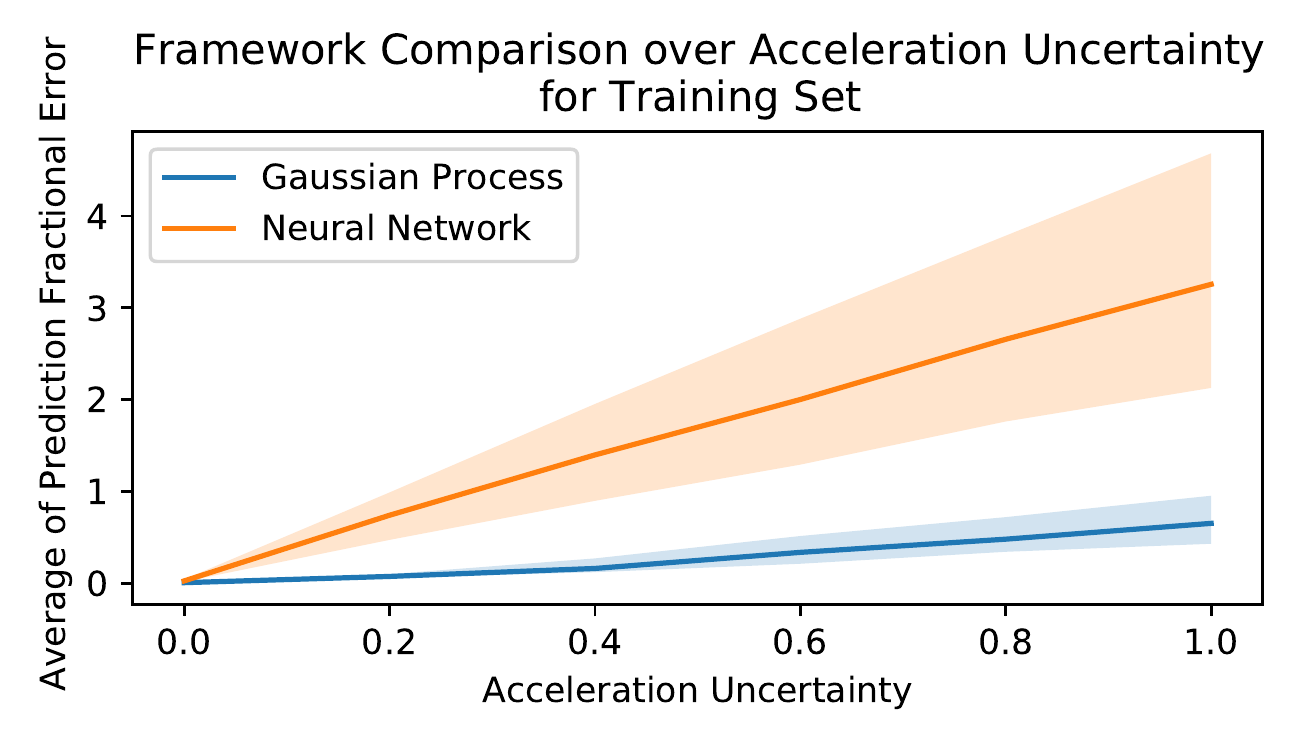}
    \includegraphics[width=0.49 \textwidth]{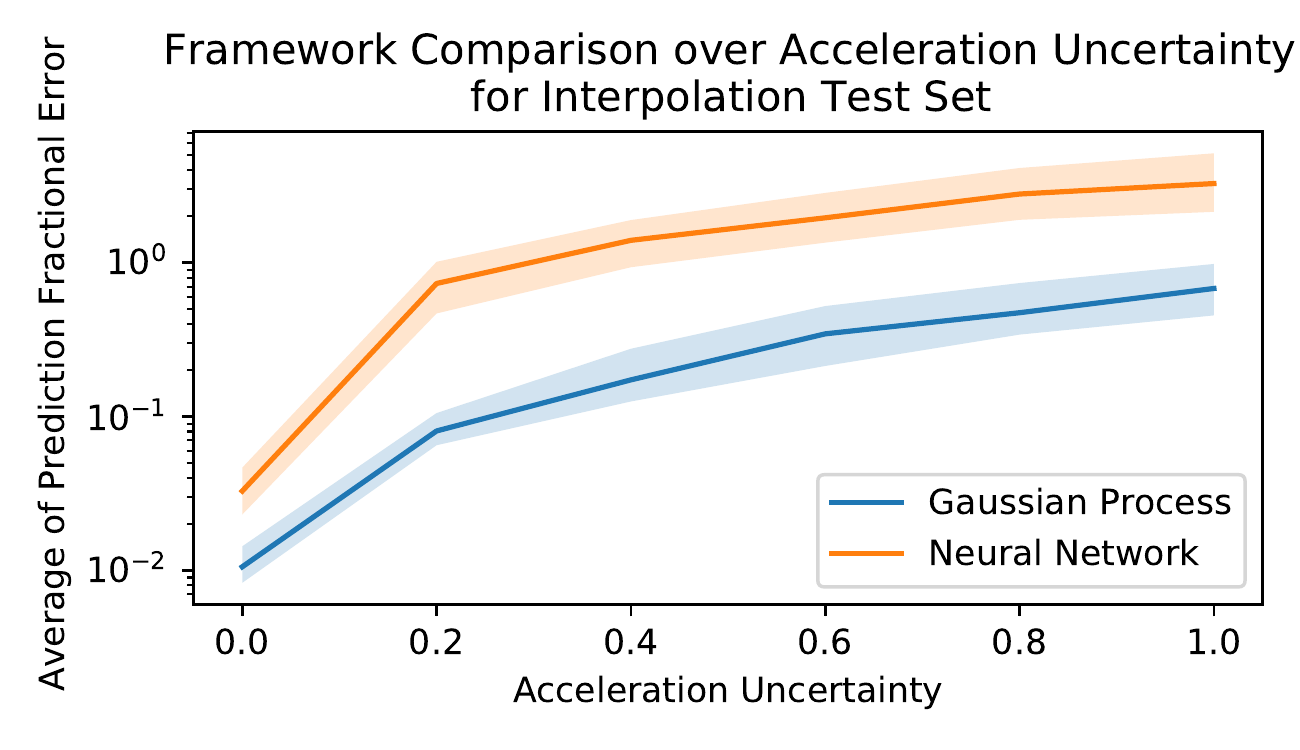}
    \includegraphics[width=0.49 \textwidth]{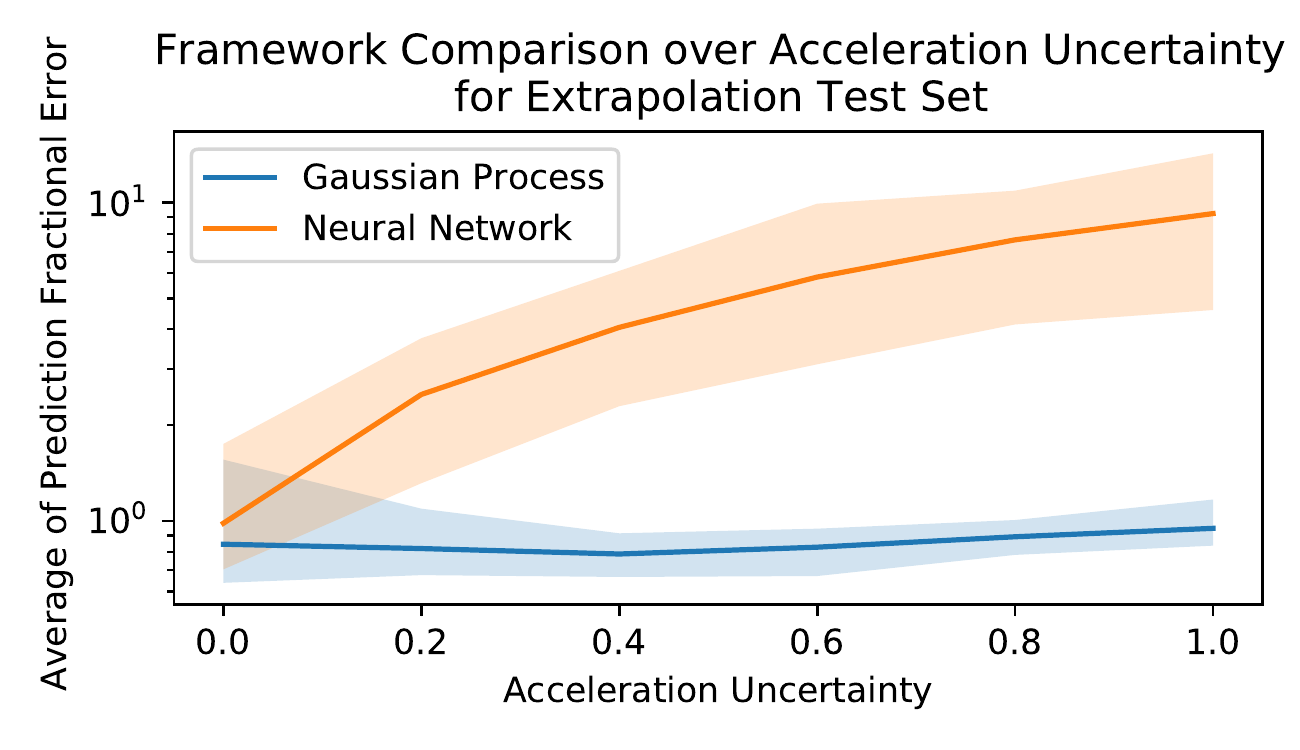}
    \caption{Comparison of Gaussian process and neural network frameworks for a range of acceleration uncertainties at low data volumes. The solid line is the median over 100 runs, and the translucent band is the interquartile range. We plot the training error on a log scale (a) for comparison at lower error and a linear scale (b) for comparison at higher error. The logarithmic scale bounds match Fig.~\ref{fig:state-uncertainty}. The training error (a-b) and interpolation test error (c) increase rapidly with the introduction of even a small acceleration uncertainty. Thereafter, the error grows roughly linearly. The extrapolation test error (d) is consistently poor throughout the range of acceleration uncertainties. However, the error increases noticeably for the neural network. In contrast, the Gaussian process error is nearly flat. At lower data volumes (as in this figure), the Gaussian process models continuously outperform the neural network models.}
    \label{fig:accel-uncertainty}
\end{figure}

Second, Fig.~\ref{fig:accel-uncertainty} details performance as the standard deviation of the acceleration uncertainty increases from 0 to 1. We fixed the state uncertainty at no noise ($\sigma_s = 0$). The training error increases much more rapidly with the introduction of acceleration uncertainty than state uncertainty. This quick jump is especially noticeable since the acceleration magnitude is often larger than the state magnitude. Once the acceleration uncertainty reaches $\sigma_a = 0.2$, the slope becomes more linear, but is already higher error than $\sigma_s = 1$ in the previous case. As before, the interpolation test error follows the training error. At $\sigma_a = 0.2$, the learned models are already less safe than the $\sigma_s = 1$ case. Following the training error, the interpolation test error flattens to roughly linear after roughly $\sigma_a = 0.2$. The trend in extrapolation test error is noticeably different for acceleration uncertainty than state uncertainty. After $\sigma_a = 0$, the Gaussian process has noticeably better performance. Indeed, neither model is robust across the range of acceleration uncertainty. The median errors for the Gaussian process (blue line in Fig.~\ref{fig:accel-uncertainty}d) are of similar magnitude to the state uncertainty case (blue line in Fig.~\ref{fig:state-uncertainty}d). However, the Gaussian process interquartile range (blue band) is tighter under acceleration uncertainty than state uncertainty. However, the neural network model cannot handle the uncertainty in acceleration error. As noted in our previous work \cite{Neamati2021NewFieldsFixed}, the spectral normalization augmentation to the neural network enforces a Lipschitz continuous output. However, the incorporation of acceleration uncertainty makes the training set discontinuous. We conclude that the Gaussian process models are safer and more robust than the neural network models for low data volumes and higher acceleration uncertainty. If we were to repeat these analyses at higher data volumes, we would arrive at a different result, as illustrated in Fig.~\ref{fig:multi-box-compare}. 

The safety and robustness of the learned models depend significantly on the data volume and uncertainty of the system. We illustrated three distinct examples with the multiple-parameter framework characterization pipeline: moderate data volumes with no uncertainty, low data volumes with state uncertainty, and low data volumes with acceleration uncertainty. As future models of learned dynamics become available, we can use the safety and robustness characterization introduced in this paper to check our refined design choices before arriving at a new, unvisited small body.

\newpage
\section{Conclusion}
In this paper, we introduce a novel method to characterize the safety and robustness of learning-based gravity models. We characterized two Gaussian process models and two neural network models on an individual level to demonstrate the breadth of characterizations provided by our method. Using the single learning framework characterization pipeline, we observed that the models are often deemed safe but are more variable in robustness. We then demonstrated full framework evaluation with the multiple-parameter framework characterization pipeline. The Gaussian processes and neural networks are comparably safe and robust in the head-to-head comparison at moderate data volumes and low uncertainty. However, numerical instabilities in the Gaussian processes yield high error outliers. The Gaussian processes are also slower than the neural networks in both training and evaluation. For comparable safety and robustness, the neural network would be more desirable than the Gaussian process for onboard spacecraft applications at moderate data volumes and low uncertainty. This recommendation reverses when operating under low data volumes and high uncertainties. The Gaussian process models consistently outperform the neural network models in safety and robustness as we increase the state or acceleration uncertainty. Robustness is consistently more difficult to achieve than safety across the tests for the full framework characterization. We recommend that future work investigate methods to incorporate robustness more directly into the learning framework. This future work includes operating on larger data sets, adding regularization specific to gravity models, and developing novel learning architectures. Working solely with simulated data, we suffer from the gap between simulation and reality. It would be most beneficial to use actual spacecraft trajectory data for characterization. In particular, we can train on the raw trajectory data and then compare the learned model predictions to the post-processed high-fidelity data. When these new learning-based gravity models are ready, we can use the safety and robustness characterization techniques introduced in this paper to evaluate our gravity models before customizing them to a new, unvisited small body.  

\bibliography{mendeley-thesis-references, thesis-bib}

\end{document}